\documentclass[lettersize,journal]{IEEEtran}
\usepackage{algorithmic}
\usepackage{algorithm}
\usepackage[caption=false,font=normalsize,labelfont=sf,textfont=sf]{subfig}
\usepackage{textcomp}
\usepackage{stfloats}
\usepackage{url}
\usepackage{verbatim}
\usepackage{graphicx}
\usepackage{cite}
\usepackage{iftex}
\usepackage{amsmath,amsfonts,amssymb,mathtools}
\ifPDFTeX
  \usepackage{newtxtext,newtxmath}
\else
  \usepackage{fontspec}
  \usepackage{unicode-math}
  \defaultfontfeatures{Ligatures=TeX}
  \renewcommand{\IEEEPARstart}[2]{#1#2}
\fi
\usepackage{array,booktabs,multirow}

\usepackage[table]{xcolor}
\usepackage{pifont}

\usepackage{float}
\usepackage{balance}
\usepackage{hyperref}

\graphicspath{{../figures/}}
\newtheorem{proposition}{Proposition}

\newtheorem{corollary}{Corollary}
\newtheorem{definition}{Definition}

\newcommand{\method}{SEARS}
\newcommand{\methodfull}{Semantic Expert Allocation with Retention-guided Synthesis}

\newsavebox{\additionaltablebox}
\newtheorem{assumption}{Assumption}[section]

\begin{document}

\title{SEARS: Semantic Expert Allocation with Retention-guided Synthesis for Multimodal Continual Instruction Tuning}

\author{
Huaijie Wang,~\IEEEmembership{} 
De Cheng,~\IEEEmembership{Senior Member,~IEEE,} 
Jie Li,~\IEEEmembership{} 
Nannan Wang,~\IEEEmembership{Senior Member,~IEEE,}
Xinbo Gao,~\IEEEmembership{Fellow,~IEEE}
\thanks{

Huaijie Wang, Jie Li, Xinbo Gao are with School of Electronic Engineering, Xidian University, Xi'an, China (email: huaijie\_wang@stu.xidian.edu.cn, leejie@mail.xidian.edu.cn, xbgao@mail.xidian.edu.cn).
De Cheng and Nannan Wang are with School of Telecommunications Engineering, Xidian University, Xi'an, China (email: dcheng@xidian.edu.cn, nnwang@xidian.edu.cn).

Corresponding Author: Xinbo Gao (xbgao@mail.xidian.edu.cn)
}
}




\maketitle

\begin{abstract}
Real-world vision--language applications require multimodal large
language models to continually adapt to new domains and instruction
requirements. 
Multimodal continual instruction tuning
must preserve both visual understanding and adherence to earlier
instruction-defined task goals and response requirements.
Expert-based adaptation can mitigate inter-task interference by
retaining task-specific capabilities in separate modules.
Instructions provide a natural cue for selecting these modules.
However, we observe that most tasks have distinct instruction
semantics, whereas some exhibit substantial overlap, making
instruction-based selection between task-specific experts ambiguous.
The challenge is therefore to organize experts according to
instruction semantics while preserving earlier functions when
experts are reused for new tasks.
To address this challenge, we propose Semantic Expert Allocation
with Retention-guided Synthesis (SEARS), which combines
instruction-guided expert allocation with post-training adaptive
synthesis. 
It retains task-specific instruction prototypes while selectively allocating or reusing projector-- Low-Rank Adaptation (LoRA) experts, termed logical roots, based on instruction similarity.
For reused roots, Adaptive Retention Synthesis (ARS) combines historical and newly adapted effective LoRA weights, incorporating new-task adaptation while limiting interference with previously learned capabilities.
Relative weight amplitudes determine the initial update proportion for each weight, and change confidence adjusts the retained fraction based on the strength of the change relative to an empirical module-wise scale. Truncated SVD
restores the prescribed rank, and ARS adds no learnable parameters or
stored importance statistics.
Experiments on four multimodal continual instruction-tuning benchmarks
demonstrate leading overall performance and improved final performance
on previously learned tasks. Code is available at \href{https://github.com/HuaijieWang/SEARS}{\textcolor{blue}{https://github.com/HuaijieWang/SEARS}}.
\end{abstract}


\section{Introduction}
\label{sec:introduction}

\IEEEPARstart{M}{ultimodal} large language models have achieved strong
performance across diverse vision--language tasks through alignment and
instruction tuning~\cite{llava2023,instructblip2023,internvl2024}, yet full
fine-tuning for downstream adaptation remains computationally and memory
intensive. Low-Rank Adaptation (LoRA)~\cite{lora2022} provides a
parameter-efficient alternative by introducing a small set of trainable
low-rank factors while keeping the pretrained backbone frozen. Owing to
its efficiency and modular design, LoRA has been widely adopted for visual
instruction tuning~\cite{smolora2025,progressivelora2025}. Nevertheless,
real-world applications require models to continually accommodate new
domains and instruction requirements without revisiting historical
training data~\cite{coin2024,hide2025,progressivelora2025}. In such
settings, successive adaptation is vulnerable to catastrophic forgetting,
as newly optimized parameters can disrupt previously learned
vision--language alignment and instruction-conditioned
responses~\cite{coin2024,smolora2025}. Consequently, multimodal continual
instruction tuning must preserve acquired capabilities (i.e., stability)
while adapting to new requirements (i.e., plasticity).

\begin{figure*}[!t]
\centering
\includegraphics[width=\textwidth]{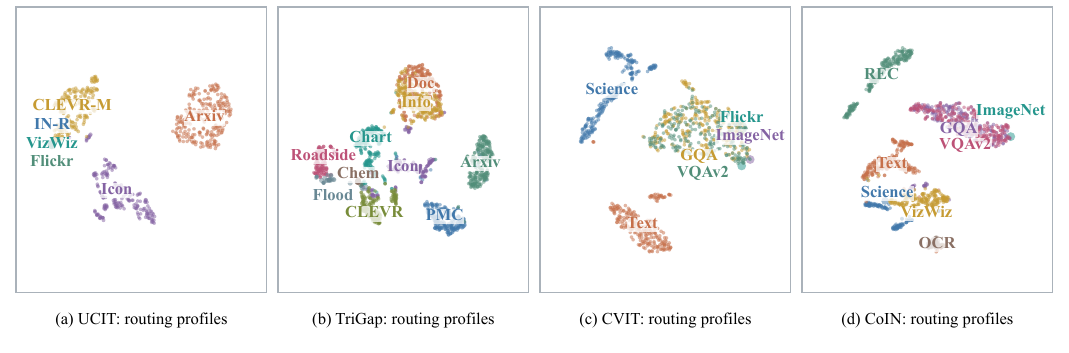}
\caption{t-SNE visualization of instruction-conditioned routing profiles
on (a) UCIT, (b) TriGap, (c) CVIT, and (d) CoIN.
Colors denote task identities.}
\label{fig:motivation}
\end{figure*}

Unlike class-incremental learning~\cite{li2017learning, zhou2024class}, which preserves recognition
of old categories under a common classification objective, \textit{multimodal
continual instruction tuning must also preserve instruction-defined
task behaviors, preventing subsequent learning from shifting responses
away from the goals and requirements of earlier
instructions}~\cite{delange2021,wangsurvey2024,coin2024,smolora2025}.
These tasks span image recognition, captioning, visual question
answering, and visual grounding, with instructions specifying both
what the model should do and how it should respond.
Forgetting may therefore also manifest as degraded instruction
alignment, with generated responses deviating from the intended
task requirements or failing to satisfy previously specified
output-format constraints~\cite{coin2024,smolora2025}.
To mitigate forgetting without repeatedly updating the full pretrained
model, parameter-efficient continual learning has explored prompt retrieval and
composition~\cite{l2p2022,dualprompt2022,codaprompt2023},
adapter organization~\cite{gae2023}, constrained low-rank
updates~\cite{olora2023,inflora2024}, and structured expert
adaptation~\cite{hide2025,clmoe2025,smolora2025,progressivelora2025}.
For approaches that maintain task-specific modules, successful
prediction also requires selecting suitable modules without task
identifiers. Instructions provide a natural cue for this selection:
they describe the requested operations and response requirements,
allowing the model to distinguish different task demands. However,
these semantic cues do not always distinguish task identities, since
different tasks may use related datasets and similar instructions.
Fig.~\ref{fig:motivation} visualizes instruction-conditioned routing
profiles across four benchmarks. Some profiles overlap, including
VizWiz and Flickr30k on UCIT, and GQA and VQAv2 on CVIT and CoIN.
When such tasks use separate experts, ambiguous instruction-based
identification can direct an example to an expert trained for a
different task. Maintaining one expert per task therefore does not
ensure reliable access to the capabilities it has acquired.

In this work, we propose \methodfull{} (\method{}), a framework that
retains a separate instruction prototype for each task and organizes
projector--LoRA experts as logical roots. Before training, instruction
similarity determines whether an arriving task reuses an existing root
or creates a new one. During training, only the selected projector and
LoRA factors are updated, while other experts remain unchanged.
Restricting updates to the selected expert reduces interference with
previously acquired instruction-dependent behaviors, helping the model
preserve its ability to satisfy earlier task objectives and response
requirements. 
For a reused root, the key challenge is that instruction similarity
supports expert sharing but does not determine how much new adaptation
should be incorporated while retaining earlier task functions.
We address this challenge with post-training Adaptive Retention
Synthesis (ARS), which combines the stored and newly trained effective
LoRA weights to regulate new adaptation and reduce interference with
earlier task-execution and response behaviors.
At inference, the model computes task probabilities from instruction
prototypes and sums the probabilities of tasks sharing an expert to
determine each expert's inference weight.

To determine this balance without accumulating additional activation or
gradient statistics, we seek a weight-based estimate of parameter
importance. Dohare et al.~\cite{dohare2024plasticity} assess connection contribution using the magnitude of the product of a unit's activation and its outgoing weight. 
Since an effective
LoRA update is a linear weighted sum of input features, its weight
magnitudes provide a proxy for relative connection contribution under a
common input. Relative amplitudes therefore propose entry-wise mixing
ratios between the stored and newly trained weights. These ratios
compare connection contributions but do not explicitly account for
current-task changes relative to the module reference. Since
acquisition-change energy is unevenly distributed, we introduce change
confidence to adaptively regulate retention of the amplitude-proposed
updates: stronger relative changes receive higher retention, whereas
weaker ones undergo greater contraction. The reference scale
is calibrated from the empirical center of these change profiles at each synthesis step. ARS combines the two signals in effective
LoRA weight space, and truncated singular value decomposition (SVD)
restores the prescribed rank. The resulting expert incorporates new adaptation without adding learnable parameters or requiring additional stored statistics
for parameter-importance estimation.

Our approach offers advantages in both theoretical analysis and empirical
performance. We derive two properties of ARS:
(1) change confidence adaptively regulates retention of the amplitude
proposal according to relative change strength;
(2) once the merging coefficients are fixed, truncated SVD minimizes the
reconstruction error of the merged weights at the prescribed rank,
retaining a compact LoRA representation.
These properties explain how synthesis controls the incorporated changes
and retains the prescribed per-expert LoRA budget. In experimental evaluations,
our method achieves leading overall performance across four multimodal
continual instruction-tuning benchmarks and improves final performance
on previously learned tasks. We summarize our contributions as follows:

\begin{itemize}
\item We propose \method{} for multimodal continual instruction tuning,
which allocates a new logical root or reuses an existing one based on
instruction similarity. It retains task-specific instruction prototypes
and aggregates task probabilities at the root level, reducing the
impact of identity ambiguity among tasks that share an expert.

\item We augment this framework with ARS for reused experts, combining relative weight amplitudes with empirically calibrated change confidence and restoring the prescribed LoRA rank through SVD. This incorporates new adaptation without
adding learnable parameters to the stored expert or requiring
additional stored statistics for parameter-importance estimation.

\item Our method achieves leading overall performance across four
multimodal continual instruction-tuning benchmarks and demonstrates
substantial improvements over existing approaches in final task
performance.
\end{itemize}

\section{Related Work}
\label{sec:related-work}

\subsection{Multimodal Large Language Models}

Multimodal large language models combine visual representations with
the language understanding and generation abilities of pretrained
LLMs~\cite{radford2021clip,grattafiori2024llama,RD-MLDG}. A central problem is
aligning visual features with the language model's input space.
BLIP-2~\cite{li2023blip2} introduces a lightweight Querying Transformer
between frozen image encoders and language models.
MiniGPT-4~\cite{zhu2024minigpt4} connects a frozen visual encoder to
a frozen language model through a trainable projection layer.
InternVL~\cite{internvl2024} scales visual pretraining and progressively
aligns the visual model with the LLM for generic visual--linguistic tasks.
These approaches establish effective connections between visual
representations and pretrained language capabilities.

Instruction tuning further teaches models to respond to diverse
natural-language requests~\cite{longpre2023flan}. LLaVA~\cite{llava2023}
constructs visual instruction-following data to train a general-purpose
visual assistant. InstructBLIP~\cite{instructblip2023} conditions visual
feature extraction on the instruction and trains across a range of
vision--language tasks. LLaVA-1.5~\cite{liu2024llava15} improves the
connector and instruction-tuning data, while
Qwen-VL~\cite{qwenvl2023} extends visual understanding to localization
and text reading. Otter~\cite{li2025otter} incorporates multimodal
in-context examples into instruction tuning to support image and video
understanding.

These advances provide a common generative interface for visual tasks
with different knowledge and response requirements. Continual adaptation
adds the requirement of learning newly arriving tasks without losing
earlier functions. Our work builds on pretrained multimodal models and
studies how lightweight adaptation states should be accessed and updated
when the complete instruction-tuning data cannot be revisited jointly.

\subsection{Parameter-Efficient Continual Learning}

Continual learning preserves knowledge under sequential training through
rehearsal, regularization, and architectural
mechanisms~\cite{delange2021,wangsurvey2024,adaptivequant2026,li2026few}. Rehearsal reinforces earlier
behavior through stored or reconstructed
examples~\cite{rebuffi2017,rolnick2019experience,buzzega2020der}, while
regularization constrains important parameters or
predictions~\cite{kirkpatrick2017,li2017learning,ekpc2026,stpr2026,symbiosis2026}. With pretrained foundation
models, recent studies implement adaptation through lightweight prompts,
adapters, and low-rank updates while keeping the backbone frozen.
According to their adaptation and preservation mechanisms, these
approaches can be grouped as follows.

1) \emph{Prompt-based methods} encode task knowledge in learnable
tokens or prefixes~\cite{lester2021prompt,li2021prefix,jia2022vpt,DPR,PADG,PAPT}.
L2P~\cite{l2p2022} and DualPrompt~\cite{dualprompt2022} retrieve prompts
from a pool, whereas CODA-Prompt~\cite{codaprompt2023} learns
input-conditioned prompt composition. The prompt pool stores task
knowledge, while input-conditioned retrieval determines which prompts
are used without requiring a supplied task identifier.

2) \emph{Adapter-based methods} introduce lightweight transformations
into pretrained models~\cite{houlsby2019,chen2022adaptformer,seca2026}.
EKPC~\cite{ekpc2026} protects shared-adapter parameters and compensates
for class-prototype drift, while StPR~\cite{stpr2026} combines selective
semantic distillation with temporal expert routing.
LAE~\cite{gae2023} accumulates adaptation through online and offline
modules, and SEMA~\cite{wang2025sema} detects distribution shifts for
selective reuse and expansion. Expert Gate~\cite{aljundi2017expert}
also maintains task-conditioned networks with expert selection.
These methods organize adaptation capacity and access to historical
expertise~\cite{he2026ckaa}. Expansion preserves existing modules at
increasing storage and selection costs, whereas reuse requires
protecting modules during subsequent updates.

3) \emph{LoRA-based methods} represent adaptation as low-rank weight
updates~\cite{lora2022,dora2024}. O-LoRA~\cite{olora2023} and
InfLoRA~\cite{inflora2024} constrain update subspaces to reduce
interference~\cite{tsd2026}. SD-LoRA~\cite{sdlora2025} decouples update
magnitudes and directions, while CL-LoRA~\cite{he2025cllora} combines
shared and task-specific adaptation. Related gradient-projection
methods~\cite{farajtabar2020ogd,gpm2021} regulate update directions
to protect historical knowledge. These approaches highlight the need
to control adaptation beyond freezing the backbone.

\subsection{Multimodal Continual Instruction Tuning}

Multimodal continual instruction tuning evaluates both the acquisition
of new task functions and the retention of earlier visual knowledge and
instruction-following behavior. CoIN~\cite{coin2024} establishes a
benchmark for this setting, identifies instruction-alignment failures,
and introduces MoELoRA. SMoLoRA~\cite{smolora2025} distinguishes
visual-understanding and instruction-following forgetting and uses
separate adaptation paths. These findings show that preserving visual
knowledge alone is insufficient if the model cannot produce the response
required by an earlier instruction. Recent approaches address this
problem through adaptation organization and constraints on continued
updates.

For adaptation organization, ModalPrompt~\cite{modalprompt2025}
selects and fuses prompts through dual-modality guidance, while
DRAPE~\cite{drape2026} generates image--instruction-conditioned prompts.
HiDe-LLaVA~\cite{hide2025} combines task-specific expansion and
task-general fusion using representation similarity.
CL-MoE~\cite{clmoe2025} employs dual routing and momentum updates,
whereas DISCO~\cite{federated2025} organizes modular adaptation.
ProgLoRA~\cite{progressivelora2025} allocates LoRA blocks by task
relevance, and Continual LLaVA~\cite{cao2024continualllava} retrieves
intrinsic and contextual increments.
PCLR~\cite{pclr2026} compresses and integrates knowledge within a
LoRA rank pool, while MoBLoRA~\cite{moblora2026} combines shared
orthogonal bases with task-specific mixing matrices.
Hyper-LLaVA~\cite{hyperllava2026} improves routing through hyperbolic
task-distribution matching and uncertainty-aware modality weighting.

For stability during continued updates, SEFE~\cite{sefe2025} addresses
response-style and knowledge forgetting, while
Octopus~\cite{liu2026octopus} uses history-free gradient orthogonalization.
KeepLoRA~\cite{keeplora2026} protects pretrained and historical
knowledge through residual-subspace adaptation, and
SiGMA~\cite{sigma2026} guides adaptation and merging through
parameter-sign consistency. $\phi$-DPO~\cite{phidpo2026} introduces
pairwise preferences and an imbalance-aware objective.
SAME~\cite{same2026} addresses both routing and expert drift through
spectral-aware routing and curvature-aware expert updates based on
historical input statistics, jointly considering capability access
and preservation.

Our work retains task-specific instruction prototypes while
selectively sharing projector--LoRA experts, termed logical roots.
Instruction similarity determines whether an arriving task creates
or reuses a root, and only the selected expert is trained.
For a reused root, ARS combines its stored and newly trained
effective LoRA weights to regulate the incorporation of new adaptation,
with fixed-rank restoration preserving the parameter budget.
At inference, task probabilities are aggregated by root to weight
the corresponding LoRA updates and projector outputs.

\section{Methodology}
\label{sec:method}

\subsection{Problem Formulation}
\label{sec:problem}

We consider multimodal continual instruction tuning over $T$ sequential
tasks. The training set of task $t$ is $\mathcal{D}_t=\{(\mathbf{X}_{t,n},\mathbf{u}_{t,n},
\mathbf{y}_{t,n})\}_{n=1}^{N_t},$
where $\mathbf{X}_{t,n}\in\mathcal{X}$ is an image,
$\mathbf{u}_{t,n}\in\mathcal{V}^{N_{t,n}^{\rm ins}}$ is its instruction,
and $\mathbf{y}_{t,n}\in\mathcal{V}^{N_{t,n}^{\rm ans}}$ is the target response.
Here $\mathcal{X}$ is the image domain and $\mathcal{V}$ is the token
vocabulary. For example $n$ in task $t$, $N_{t,n}^{\rm ins}$ and
$N_{t,n}^{\rm ans}$ count the instruction and target-response tokens,
respectively.
Only $\mathcal{D}_t$ is accessible at stage $t$; earlier training sets
cannot be revisited. At inference, the model must answer instructions
from all observed tasks without their task identifiers. The objective
is to acquire the arriving task while retaining earlier visual knowledge
and instruction-dependent responses within one generative model.
Unlike standard class-incremental learning under a common classification
objective, this setting must retain the ability to execute earlier
instruction-defined tasks and satisfy their response
requirements.

\subsection{Overall Framework}
\label{sec:overview}

\begin{figure*}[!t]
\centering
\includegraphics[width=\textwidth]{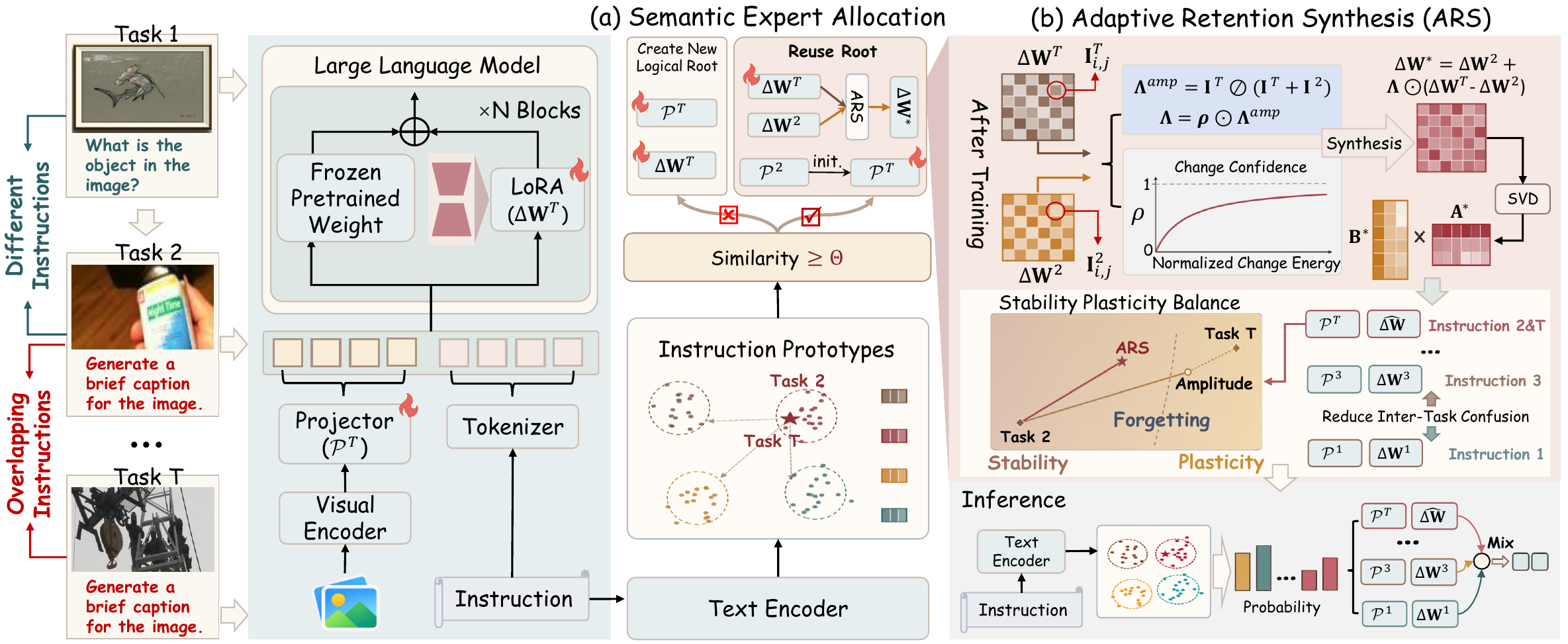}
\caption{
Overview of \method{}.
(a) Semantic Expert Allocation compares instruction prototypes before
training. In this example, Tasks~2 and~$T$ share a logical root,
while others retain separate roots.
(b) After training, ARS uses relative weight amplitudes and change
confidence to balance retention and acquisition in the reused root.
Truncated SVD restores the LoRA rank, and the trained projector is
retained directly. Darker matrix cells indicate larger absolute weights.
At inference, task probabilities are summed by root and used to weight effective LoRA updates and projector outputs.
}
\label{fig:framework}
\end{figure*}

The framework of \method{} is illustrated in Fig.~\ref{fig:framework}.
For task $t$, the visual encoder extracts token features
$\mathbf{X}^{\rm vis}$ from image $\mathbf{X}_{t,n}$. The projector
$\mathcal{P}_g(\cdot;\boldsymbol{\Phi}_g)$, with parameter vector
$\boldsymbol{\Phi}_g$, maps these features into the language model's
input space. The projected features and instruction tokens
$\mathbf{u}_{t,n}$ are used to generate a response autoregressively.
The projector and LoRA factor pairs
$\{\mathbf{A}_{g,\ell},\mathbf{B}_{g,\ell}\}_{\ell=1}^{L}$
form an expert indexed by logical root $g$, where $\ell$ indexes
an adapted mapping and $L$ is the number of adapted mappings.
A frozen instruction encoder $\mathcal{F}_{\rm txt}$ produces
$D_c$-dimensional features, whose normalized means form task prototypes.
Before training, Semantic Expert Allocation compares the current
prototype $\mathbf{c}^{\rm scan}_t\in\mathbb{R}^{D_c}$ with stored
prototypes $\{\mathbf{c}^{\rm store}_j\}_{j<t}$ to select an existing
root $g_t$ or create a new one. The superscripts ${\rm scan}$ and
${\rm store}$ distinguish the prototype used before acquisition from
the one retained afterward. Each task keeps its own prototype even when
several tasks share a root. For reuse, the selected root's stored
projector parameters and LoRA factors initialize current-task training,
and its pre-training effective LoRA weights are retained for subsequent
synthesis. 
When the semantic similarity criterion for reuse is not met,
we initialize the new root's projector from the pretrained
multimodal projector, initialize its LoRA factor $\mathbf{A}_{g_t,\ell}$
with Gaussian random weights,
and initialize $\mathbf{B}_{g_t,\ell}$ to zero.
Only the selected projector parameters
$\boldsymbol{\Phi}_{g_t}$ and LoRA factors
$\{\mathbf{A}_{g_t,\ell},\mathbf{B}_{g_t,\ell}\}_{\ell=1}^{L}$
are optimized on $\mathcal{D}_t$; the visual encoder, pretrained
language-model weights, and other experts remain frozen.

After training, a newly created root is stored directly, while a reused
root is updated by ARS. As illustrated in Fig.~\ref{fig:framework}(b),
ARS starts from the stored effective LoRA weights and uses relative
weight amplitudes to propose an entry-wise update toward the newly
trained weights. Change confidence then adjusts the fraction of the
proposed update retained at each entry: changes that are stronger
relative to the module's reference scale retain a larger fraction,
whereas weaker changes leave the weights closer to their stored values.
This balances the preservation of earlier capabilities with adaptation
to new instructions. Truncated SVD restores the prescribed rank
$R_\ell$, and the resulting LoRA factors are stored with the newly
trained projector. At inference, task probabilities are summed by root
to weight the effective LoRA updates and projector outputs.

\subsection{Preliminaries}
\label{sec:preliminaries}

\paragraph{Low-Rank Adaptation}
For adapted mapping $\ell\in\{1,\ldots,L\}$ with frozen weight
$\mathbf{W}_{0,\ell}\in
\mathbb{R}^{D_{{\rm out},\ell}\times D_{{\rm in},\ell}}$,
LoRA~\cite{lora2022} introduces the effective update
\begin{equation}
\Delta\mathbf{W}_{g,\ell}
=\varsigma_\ell\mathbf{B}_{g,\ell}\mathbf{A}_{g,\ell},
\label{eq:lora}
\end{equation}
where $\mathbf{A}_{g,\ell}\in
\mathbb{R}^{R_\ell\times D_{{\rm in},\ell}}$,
$\mathbf{B}_{g,\ell}\in
\mathbb{R}^{D_{{\rm out},\ell}\times R_\ell}$,
$R_\ell$ is the prescribed rank, and $\varsigma_\ell>0$ is the fixed
execution scale. The operator acts on a module input
$\mathbf{h}_\ell\in\mathbb{R}^{D_{{\rm in},\ell}}$ and has the same
shape as $\mathbf{W}_{0,\ell}$. All roots use this common frozen
backbone; their effective LoRA weights are adaptations relative to it.

\begin{definition}[Logical root]
\label{def:logical-root}
A logical root $g$ is an executable expert consisting of a multimodal
projector $\mathcal{P}_g$ and the LoRA factor pairs
$\{\mathbf{A}_{g,\ell},\mathbf{B}_{g,\ell}\}_{\ell=1}^{L}$.
Its projector parameters are collected in the vector
$\boldsymbol{\Phi}_g\in\mathbb{R}^{D_{\rm proj}}$ by vectorizing its
matrices and biases, where $D_{\rm proj}$ is their total parameter
count. Concatenating this vector and the vectorized LoRA factors gives
$\boldsymbol{\theta}_g\in\mathbb{R}^{D_{\rm proj}+D_{\rm LoRA}}$, with
$D_{\rm LoRA}=\sum_{\ell=1}^{L}R_\ell
(D_{{\rm in},\ell}+D_{{\rm out},\ell})$.
The assignment $\mathcal{H}_t:\{1,\ldots,t\}\rightarrow\mathcal{G}_t$
maps each observed task identity to a root, where
$\mathcal{G}_t\subseteq\{1,\ldots,t\}$ indexes the tasks that created
the retained experts. Tasks assigned to one root share its executable
parameters but retain separate instruction prototypes.
\end{definition}

\paragraph{Inter- and Intra-Task Performance}
\label{sec:performance-analysis}

Continual instruction performance depends on both access to suitable
expertise and the capabilities retained in that expertise. To express
this relationship, let $\xi=(\mathbf{X},\mathbf{u},\mathbf{y})$ be an
evaluation example and let $\mathcal{S}$ denote the event that the generated response meets
its task's success criterion.
Let $J$ be the true task identity and
$\widehat J=\arg\max_j\pi_j^{\rm task}(\mathbf{u})$ the identity with
the largest routing weight, with ties resolved by a fixed rule.
Here $\pi_j^{\rm task}(\mathbf{u})$ is defined in
Eq.~(\ref{eq:identity-router}); $\Pr$ denotes probability over evaluation
examples and their responses. 
We write $\mathcal{C}=\{\mathcal{H}_t(\widehat J)=\mathcal{H}_t(J)\}$
for the event of correct root association and $\mathcal{C}^{c}$ for
its complement. The law of total probability gives
\begin{equation}
\begin{aligned}
\Pr(\mathcal{S})={}&
\Pr(\mathcal{C})\Pr(\mathcal{S}\mid\mathcal{C})\\
&+\Pr(\mathcal{C}^{c})\Pr(\mathcal{S}\mid\mathcal{C}^{c}).
\end{aligned}
\label{eq:performance-decomposition}
\end{equation}
Equation~(\ref{eq:performance-decomposition}) separates the frequency
of correct root association from response success under each
association outcome. In both cases, the response is generated with
the soft mixture in Section~\ref{sec:compact}.
With shared roots, an incorrect identity prediction can still refer to
the same expert. For a fixed assignment, correct identity prediction
implies correct root association, yielding
$\Pr(\mathcal{C})\geq\Pr(\widehat J=J)$.
During soft execution, redistributing identity probabilities within a
root preserves its contribution whenever their total mass is unchanged.
These relationships motivate Semantic Expert Allocation for accessing
acquired expertise and ARS for preserving that expertise during
continued adaptation.
Appendix.\textcolor{red}{B-A} provides the detailed analysis.

\subsection{Semantic Expert Allocation}
\label{sec:module-i}
The preceding analysis examines continual performance through both
expert matching across tasks and task competence within experts.
Instructions describe task objectives and response requirements,
providing a natural basis for selecting appropriate experts.
However, different tasks may have similar instruction requirements,
so assigning separate experts solely by task identity does not
reflect their semantic relationships. We therefore use instruction
similarity to determine whether an arriving task reuses an existing
logical root or creates a new one.
A frozen instruction encoder $\mathcal{F}_{\rm txt}$~\cite{radford2021clip}
maps instructions into a common semantic space. Before training, we scan
$N_t^{\rm scan}$ current instructions, denoted by
$\{\mathbf{u}_{t,n}^{\rm scan}\}_{n=1}^{N_t^{\rm scan}}$, and form
\begin{equation}
\mathbf{c}^{\rm scan}_t
=\operatorname{Norm}\!\left(
\frac{1}{N_t^{\rm scan}}\sum_{n=1}^{N_t^{\rm scan}}
\mathcal{F}_{\rm txt}(\mathbf{u}^{\rm scan}_{t,n})\right),
\label{eq:prototype}
\end{equation}
where $\operatorname{Norm}(\mathbf{v})=\mathbf{v}/\|\mathbf{v}\|_2$
for any nonzero vector $\mathbf{v}$. For $t>1$, the closest
historical identity is
\begin{equation}
s_{t,j}=(\mathbf{c}^{\rm scan}_t)^\top\mathbf{c}^{\rm store}_j,
\qquad j_t^\star=\arg\max_{j<t}s_{t,j}.
\label{eq:similarity}
\end{equation}
Given the reuse threshold $\Theta$, the root decision is
\begin{equation}
g_t=\begin{cases}
\mathcal{H}_{t-1}(j_t^\star),
&t>1,\ s_{t,j_t^\star}\geq\Theta,\\
t,&\text{otherwise}.
\end{cases}
\label{eq:gate}
\end{equation}
When $t>1$ and $s_{t,j_t^\star}\geq\Theta$, task $t$ reuses the existing
root $g_t=\mathcal{H}_{t-1}(j_t^\star)$. The projector parameters
$\boldsymbol{\Phi}_{g_t}$ and LoRA factors
$\{\mathbf{A}_{g_t,\ell},\mathbf{B}_{g_t,\ell}\}_{\ell=1}^{L}$
are initialized from that root's latest stored state.
Otherwise, a new root $g_t=t$ is created.
We set $\mathcal{H}_t(t)=g_t$ while keeping all historical
assignments unchanged.
During acquisition, we optimize only $\boldsymbol{\theta}_{g_t}$.
Let $y_{t,n,q}$ be target token $q$ in $\mathbf{y}_{t,n}$ and
$\mathbf{y}_{t,n,<q}$ its preceding target tokens.
Let $p_{\mathcal M}$ denote the conditional next-token distribution of
the multimodal model $\mathcal M$. The mean negative
log-likelihood over all target-response tokens is
\begin{equation}
\begin{aligned}
\mathcal{L}_t(\boldsymbol{\theta}_{g_t})=-&
\frac{1}{\sum_{n=1}^{N_t}N_{t,n}^{\rm ans}}
\sum_{n=1}^{N_t}\sum_{q=1}^{N_{t,n}^{\rm ans}}\\
&\log p_{\mathcal M}\!\left(
y_{t,n,q}\mid\mathbf{X}_{t,n},\mathbf{u}_{t,n},
\mathbf{y}_{t,n,<q};\boldsymbol{\theta}_{g_t}\right).
\end{aligned}
\label{eq:training-loss}
\end{equation}

The backbone and all other roots remain frozen. We accumulate the
instruction features encountered during this training and store their
normalized mean as $\mathbf{c}^{\rm store}_t$. 
For a newly created root, we directly store the trained projector
and LoRA factors. When a root is reused, training continues from its
stored state, with the pre-training and acquired effective LoRA weights
denoted by $\Delta\mathbf{W}^{\rm old}_{g_t,\ell}$ and
$\Delta\mathbf{W}^{\rm new}_{g_t,\ell}$, respectively.
ARS combines these two effective linear operators, while the projector
is handled separately because parameter interpolation in its nonlinear
two-layer mapping does not generally correspond to output interpolation.
For an interpolation coefficient $\gamma\in(0,1)$,
\begin{equation}
\begin{aligned}
&\mathcal{P}_{g_t}\!\left(
\mathbf{X}^{\rm vis};
(1-\gamma)\boldsymbol{\Phi}^{\rm old}_{g_t}
+\gamma\boldsymbol{\Phi}^{\rm new}_{g_t}
\right)\\
&\quad\not\equiv
(1-\gamma)\mathcal{P}_{g_t}\!\left(
\mathbf{X}^{\rm vis};\boldsymbol{\Phi}^{\rm old}_{g_t}
\right)+
\gamma\mathcal{P}_{g_t}\!\left(
\mathbf{X}^{\rm vis};\boldsymbol{\Phi}^{\rm new}_{g_t}
\right),
\end{aligned}
\label{eq:projector-interpolation}
\end{equation}
where $\not\equiv$ denotes nonidentity. We retain the newly trained projector for current-task visual--language alignment
and apply ARS to the effective LoRA weights.
Appendix.\textcolor{red}{B-E} examines continued projector adaptation
together with LoRA synthesis and root-level inference through paired
comparisons.

Each accepted reuse avoids allocating an additional expert while
preserving the arriving identity's prototype. The root count is
\begin{equation}
K_t=|\mathcal{G}_t|=t-\sum_{u=2}^{t}\iota_u,
\label{eq:root-count}
\end{equation}
where $\iota_u=\mathbb I[g_u\ne u]$ indicates accepted reuse for
task $u$. The resulting state budget
depends on which tasks share experts.

\subsection{Adaptive Retention Synthesis}
\label{sec:module-ii}

Semantic expert allocation allows related tasks to share an expert,
but continued training may interfere with that expert's previously
learned functions. Moreover, current-task adaptation changes different
connections to different extents, which a uniform mixing ratio does
not capture. We therefore introduce ARS to regulate the incorporation
of new adaptation at each weight entry while keeping the stored
expert compact. Its synthesis coefficients are computed directly
from the stored and newly trained effective LoRA weights, avoiding
persistent activation or gradient-based importance statistics.
Relative amplitude provides an initial balance between the two states,
while change confidence regulates update retention according to
change strength relative to an empirically calibrated module reference.
Truncated SVD then restores the prescribed rank without adding
parameters to the stored expert.

\subsubsection{Relative Amplitude and Empirical Calibration}
Inspired by the connection utility of Dohare et
al.~\cite{dohare2024plasticity}, we seek an importance proxy that requires no additional activation or gradient statistics. In a linear mapping, connection contribution depends on both the input activation and its weight. Our statistics show nearly identical activation across
related tasks. When the activation statistics agree, their common factor cancels from the relative connection contribution. We therefore use weight magnitudes as an importance proxy and adopt the squared-energy form for synthesis. Appendix.\textcolor{red}{B-B} provides the detailed statistics and approximation analysis.
Suppress root and module indices
and define
\begin{equation}
\mathbf{D}=\Delta\mathbf{W}^{\rm new}-\Delta\mathbf{W}^{\rm old},
\qquad\mathbf{Z}=\mathbf{D}^{\odot2}.
\label{eq:change}
\end{equation}
The amplitude importance and proposed new-state share are
\begin{equation}
\begin{aligned}
&\mathbf{I}^{\zeta}
=(|\Delta\mathbf{W}^{\zeta}|+\varepsilon\mathbf{1})^{\odot\alpha},
\quad \zeta\in\{\mathrm{old},\mathrm{new}\},\\
&\boldsymbol{\Lambda}^{\rm amp}
=\mathbf{I}^{\rm new}\oslash
(\mathbf{I}^{\rm old}+\mathbf{I}^{\rm new}+\varepsilon\mathbf{1}),
\end{aligned}
\label{eq:amp-share}
\end{equation}
where $\alpha=2$, $\varepsilon>0$ is a numerical constant, $\mathbf{1}$
is an all-ones matrix of the same size as $\mathbf{D}$, and $\odot$ and $\oslash$ denote
element-wise product and division. 
The resulting share favors the acquired state where the magnitude of its effective weight is larger relative to the stored state.
For mapping $\ell$, the importance and coefficient matrices, together
with $\mathbf D$ and $\mathbf Z$, lie in
$\mathbb R^{D_{{\rm out},\ell}\times D_{{\rm in},\ell}}$.

The effective weight is the product of the LoRA factors, including its
execution scale. 
For any invertible $\mathbf{C}$, the pairs
$(\mathbf{B},\mathbf{A})$ and
$(\mathbf{B}\mathbf{C},\mathbf{C}^{-1}\mathbf{A})$ have the same product.
We therefore compute amplitude and change statistics from the
effective LoRA weight matrices. Equivalent factorizations of the stored and acquired
weights consequently yield the same synthesis coefficients.
Let $\mathcal{U}_t$ index the participating mappings of the reused root.
For mapping $\ell\in\mathcal{U}_t$, let
$Z_{t,\ell,ab}=(\mathbf{Z}_{t,\ell})_{ab}$ be the squared change in
row $a$ and column $b$. We compute
\begin{equation}
\begin{aligned}
m_{t,\ell}&=\operatorname{Med}_{\rm lo}\!\left(\{Z_{t,\ell,ab}\}_{a,b}\right),\\
\bar z_{t,\ell}&=\frac{1}{n_\ell}\sum_{a,b}Z_{t,\ell,ab}.
\end{aligned}
\label{eq:module-change-statistics}
\end{equation}
where $n_\ell=D_{{\rm out},\ell}D_{{\rm in},\ell}$ is the module's
entry count. The operator $\operatorname{Med}_{\rm lo}$ returns the
middle entry after sorting, using the lower of the two central entries
for an even count. Over mappings with positive median and mean,
$\mathcal{U}_t^{+}=\{\ell\in\mathcal{U}_t:m_{t,\ell}>0,\bar z_{t,\ell}>0\}$,
we define
\begin{equation}
\eta_{t,\ell}=\frac{m_{t,\ell}}{\bar z_{t,\ell}},
\qquad
\kappa_t=\operatorname{Med}_{\rm mid}
\{\eta_{t,\ell}:\ell\in\mathcal{U}_t^{+}\}.
\label{eq:empirical-kappa}
\end{equation}
The operator $\operatorname{Med}_{\rm mid}$ returns the middle ratio
for an odd count and averages the two central ratios for an even count.
If $\mathcal{U}_t^{+}$ is empty, we keep the stored LoRA factors and
the newly trained projector.
The ratio $\eta_{t,\ell}$ is unchanged when a module's
displacement is multiplied by a positive constant. Thus the center
compares change profiles independently of their absolute scales, with
equal weight assigned to each module.

Pooling all entries into one distribution would weight modules according
to their entry counts. The present construction instead obtains a ratio
within each module and then estimates a shared center across modules.
Attention and feed-forward mappings therefore contribute one ratio
each, while retaining their own absolute change scales. The calibration
separates the common reference statistic from the module-specific magnitude
used by the gate. The module reference is then
\begin{equation}
\tau_{t,\ell}^{2}
=\frac{m_{t,\ell}}{\kappa_t}.
\label{eq:scale}
\end{equation}
For $\ell\in\mathcal{U}_t^{+}$, this reference can also be written as
$(\eta_{t,\ell}/\kappa_t)\bar z_{t,\ell}$. A mapping whose ratio equals
the empirical center thus uses its own mean squared change as the
reference. The superscript in $\tau_{t,\ell}^{2}$ denotes a scale in
squared-weight units.

\subsubsection{Change-Dependent Retention}
For the current synthesis step, suppress the index $t$ and write
$\mathbf{D}_\ell^{\rm amp}
=\boldsymbol{\Lambda}^{\rm amp}_\ell\odot\mathbf{D}_\ell$ for the
amplitude-proposed update. Set
$\Omega_\ell=\{(a,b):Z_{\ell,ab}>0\}$ and
$\chi_\ell=\max\{\tau_\ell^2,\varepsilon\}+\varepsilon$.
The retained update solves
\begin{equation}
\begin{aligned}
\mathbf{D}_\ell^{\rm ret}=\arg\min_{\mathbf{Y}}\quad
&\frac12\|\mathbf{Y}-\mathbf{D}_\ell^{\rm amp}\|_F^2
+\frac12\sum_{(a,b)\in\Omega_\ell}
\frac{\chi_\ell}{Z_{\ell,ab}}Y_{ab}^2,\\
&Y_{ab}=0\quad\text{for }(a,b)\notin\Omega_\ell.
\end{aligned}
\label{eq:confidence-objective}
\end{equation}
The penalty $\chi_\ell/Z_{\ell,ab}$ is stronger for weaker relative
changes. The unique solution is
$\mathbf{D}_\ell^{\rm ret}
=\boldsymbol{\rho}_\ell\odot\mathbf{D}_\ell^{\rm amp}$, where
\begin{equation}
\rho_{\ell,ab}=\frac{Z_{\ell,ab}}{Z_{\ell,ab}+\chi_\ell}.
\label{eq:change-score}
\end{equation}
The entries $\rho_{\ell,ab}$ form the confidence matrix
$\boldsymbol{\rho}_\ell\in
\mathbb{R}^{D_{{\rm out},\ell}\times D_{{\rm in},\ell}}$.
At a fixed reference, stronger changes retain a larger fraction of the
proposed update; unchanged entries have zero retention.
Appendix.\textcolor{red}{B-D} derives this rule, and
Appendix.\textcolor{red}{B-E} evaluates its effect on new- and
earlier-task responses.

The resulting coefficient and dense target are
\begin{equation}
\begin{aligned}
\boldsymbol{\Lambda}&=\boldsymbol{\rho}\odot\boldsymbol{\Lambda}^{\rm amp},\\
\Delta\mathbf{W}^{\star}&=\Delta\mathbf{W}^{\rm old}
+\boldsymbol{\Lambda}\odot\mathbf{D}.
\end{aligned}
\label{eq:interpolation}
\end{equation}
\subsubsection{Fixed-Rank Reconstruction and Local Stability}
\label{sec:rank-restoration}

Element-wise synthesis can increase matrix rank. We reconstruct the
dense target under the prescribed LoRA rank budget.

For the computed synthesis coefficients $\boldsymbol{\Lambda}$, we
restore the prescribed rank by minimizing weighted deviations from the
stored and acquired effective weights:
\begin{equation}
\begin{aligned}
\mathcal{J}(\mathbf{M})
={}&\frac12\left\|
\sqrt{\mathbf{1}-\boldsymbol{\Lambda}}\odot
(\mathbf{M}-\Delta\mathbf{W}^{\rm old})\right\|_F^2\\
&+\frac12\left\|
\sqrt{\boldsymbol{\Lambda}}\odot
(\mathbf{M}-\Delta\mathbf{W}^{\rm new})\right\|_F^2.
\end{aligned}
\label{eq:weighted-objective}
\end{equation}
Let
$\mathbf{M}\in\mathbb{R}^{D_{\mathrm{out},\ell}\times D_{\mathrm{in},\ell}}$
denote a candidate effective LoRA weight update. The complementary weights specify how closely each entry should follow
the previous and newly trained weights. Because they sum to one, this objective
reduces to a Frobenius reconstruction of $\Delta\mathbf{W}^{\star}$.

\begin{corollary}[Fixed-rank reconstruction]
\label{thm:synthesis}
Let the stored and acquired effective weights be finite matrices of the
same shape. Hold $\boldsymbol{\Lambda}$ fixed with entries in $[0,1]$.
The unique dense minimizer of $\mathcal{J}$ is
$\Delta\mathbf{W}^{\star}$. If
$\Delta\mathbf{W}^{\star}=\mathbf{U}\boldsymbol{\Sigma}\mathbf{V}^{\top}$,
then
\begin{equation}
\widehat{\Delta\mathbf{W}}
=\mathbf{U}_{\bar R}\boldsymbol{\Sigma}_{\bar R}\mathbf{V}_{\bar R}^{\top}
\in\underset{\operatorname{rank}(\mathbf{M})\leq\bar R}{\arg\min}
\mathcal{J}(\mathbf{M}),
\label{eq:rank-weighted-optimum}
\end{equation}
where $d_{\min}=\min(D_{\rm in},D_{\rm out})$ and
$\bar R=\min(R,d_{\min})$. In this compact SVD,
$\mathbf{U}\in\mathbb{R}^{D_{\rm out}\times d_{\min}}$,
$\boldsymbol{\Sigma}\in\mathbb{R}^{d_{\min}\times d_{\min}}$, and
$\mathbf{V}\in\mathbb{R}^{D_{\rm in}\times d_{\min}}$.
Let $\sigma_i=(\boldsymbol{\Sigma})_{ii}$ be the $i$th singular value
in nonincreasing order. The subscript $\bar R$ indicates that the
first $\bar R$ singular triplets are retained. The reconstruction
residual norm satisfies $e_R^2=\sum_{i>\bar R}\sigma_i^2$, and
\begin{equation}
\begin{aligned}
\|\boldsymbol{\Lambda}\odot\mathbf{D}\|_F
&\leq\|\boldsymbol{\Lambda}^{\rm amp}\odot\mathbf{D}\|_F,\\
\|\widehat{\Delta\mathbf{W}}-\Delta\mathbf{W}^{\rm old}\|_F
&\leq\|\boldsymbol{\Lambda}\odot\mathbf{D}\|_F+e_R.
\end{aligned}
\label{eq:post-rank-bound}
\end{equation}
\end{corollary}

Equation~(\ref{eq:post-rank-bound}) further relates the
restored LoRA displacement to the retained update and truncation
residual. 
Appendix.\textcolor{red}{B-C} derives the fixed-rank optimum and
scale-preserving recovery of LoRA factors.
\begin{proposition}[Local historical response-loss bound]
\label{prop:historical-risk}
Let $\boldsymbol{\vartheta}$ collect the reused root's vectorized
effective LoRA matrices and projector parameters. With other roots,
task prototypes, and inference weights fixed, let $\mathcal{R}_{\rm old}$
be the expected response loss on earlier tasks.
Assume its gradient is $\beta$-Lipschitz along the segment from the
stored effective state $\boldsymbol{\vartheta}_o$ to the final effective state $\boldsymbol{\vartheta}_f$.
Set $b_\ell=\|\boldsymbol{\Lambda}_\ell\odot\mathbf{D}_\ell\|_F+e_{R,\ell}$,
$d_\Phi=\|\boldsymbol{\Phi}^{\rm new}-\boldsymbol{\Phi}^{\rm old}\|_2$,
and let $G_\ell,G_\Phi$ be the corresponding block-gradient norms at
$\boldsymbol{\vartheta}_o$. The signed loss difference
$\Delta\mathcal{R}_{\rm old}:=\mathcal{R}_{\rm old}(\boldsymbol{\vartheta}_f)
-\mathcal{R}_{\rm old}(\boldsymbol{\vartheta}_o)$ satisfies
\begin{equation}
\Delta\mathcal{R}_{\rm old}
\leq\sum_\ell\left(G_\ell b_\ell+\frac{\beta}{2}b_\ell^2\right)
+G_\Phi d_\Phi+\frac{\beta}{2}d_\Phi^2.
\label{eq:historical-risk-bound}
\end{equation}
\end{proposition}
The retained LoRA update and reconstruction residual enter the bound
through $b_\ell$, while $d_\Phi$ captures projector adaptation.
Appendix.\textcolor{red}{B-D} provides the response-loss analysis,
including changes in root probabilities.

\subsection{Inference}
\label{sec:compact}

After stage $t$, the $E=t$ task identities are assigned to $K=K_t$
distinct roots, enumerated as $\mathcal{G}_t=\{r_1,\ldots,r_K\}$.
Here $k$ is a position in this enumeration, and $r_k$ is the
corresponding root identifier. For instruction $\mathbf{u}$, let
$\mathbf{f}(\mathbf u)
=\operatorname{Norm}(\mathcal{F}_{\rm txt}(\mathbf u))
\in\mathbb{R}^{D_c}$ and compute
\begin{equation}
\begin{aligned}
s_j(\mathbf u)&=\mathbf{f}(\mathbf u)^\top\mathbf{c}^{\rm store}_j,\\
\boldsymbol{\pi}^{\rm task}(\mathbf u)
&=\operatorname{softmax}\bigl(\mathbf{s}(\mathbf u)/T_r\bigr).
\end{aligned}
\label{eq:identity-router}
\end{equation}
Here $\mathbf{s}(\mathbf u)\in\mathbb{R}^{E}$ collects the cosine
scores, $\pi_j^{\rm task}(\mathbf u)$ is the routing weight of
identity $j$, and $T_r>0$ is the routing temperature. Define the
root-membership matrix $\mathbf{G}_t\in\{0,1\}^{E\times K}$ by
\begin{equation}
\begin{aligned}
G_{t,jk}&=\mathbb{I}[\mathcal{H}_t(j)=r_k],\\
\boldsymbol{\pi}^{\rm root}(\mathbf u)
&=\mathbf{G}_t^\top\boldsymbol{\pi}^{\rm task}(\mathbf u).
\end{aligned}
\label{eq:root-router}
\end{equation}
In $\boldsymbol{\pi}^{\rm root}(\mathbf u)\in\mathbb R^K$, entry
$\pi_k^{\rm root}(\mathbf u)$ is the routing weight of root $r_k$.
Each identity belongs to one root, so the root weights
are nonnegative and sum to one. Both adaptation paths use these weights:
\begin{align}
&\Delta\mathbf{W}^{\rm route}_{\ell}
=\sum_{k=1}^{K}\pi_k^{\rm root}(\mathbf u)\Delta\mathbf{W}_{r_k,\ell},
\label{eq:routed-lora}\\
&\mathbf{O}^{\rm soft}
=\sum_{k=1}^{K}\pi_k^{\rm root}(\mathbf u)
\mathcal{P}_{r_k}(\mathbf{X}^{\rm vis};\boldsymbol{\Phi}_{r_k}).
\label{eq:routed-projector}
\end{align}
Here $\mathbf{X}^{\rm vis}\in\mathbb{R}^{N_v\times D_v}$ contains
$N_v$ visual tokens from the frozen visual encoder, each with feature
dimension $D_v$. Each projector maps these tokens to the language
model's input space of dimension $D_m$, so its output and the
weighted mixture $\mathbf{O}^{\rm soft}$ both lie in
$\mathbb{R}^{N_v\times D_m}$. 
Projectors are
mixed after their individual nonlinear transformations; LoRA updates
are mixed as effective linear operators.

The same grouping allows the bank to store only reachable roots.
For $D_{\rm LoRA}=\sum_\ell R_\ell(D_{{\rm in},\ell}+D_{{\rm out},\ell})$
LoRA entries and $D_{\rm proj}$ projector entries per root, the saving
relative to an identity-wise bank is
\begin{equation}
\Delta D_{\rm bank}=(E-K)(D_{\rm LoRA}+D_{\rm proj}).
\label{eq:persistent-saving}
\end{equation}
Thus, semantic sharing controls the number of states and SVD controls
their individual rank, while root aggregation preserves the soft
operators.

\begin{table*}[t]
\centering
\caption{Comparison with existing methods on UCIT.
The best results are in bold, and the second-best results are underlined.}
\setlength{\tabcolsep}{3.3pt}
\renewcommand{\arraystretch}{1.08}
\small
\begin{tabular}{lccccccc|c>{\columncolor{gray!10}}c>{\columncolor{gray!10}}cc}
\toprule
Method & Venue & ImageNet-R &  ArXivQA & VizWiz & IconQA & CLEVR-Math & Flickr30k  & MFT & MFN & MAA & BWT \\
\hline
Zero-shot & --
& 16.27 & 53.73 & 38.39 & 19.20 & 20.63 & 41.88
& 31.68 & -- & -- & -- \\
Individual & --
& 91.67 & 90.83 & 57.87 & 78.43 & 76.63 & 61.72
& 76.19 & -- & -- & -- \\
\hline
LoRA-FT~\cite{lora2022} & ICLR 2022
& 58.03 & 77.63 & 44.39 & 67.40 & 61.77 & \textbf{58.22}
& \textbf{76.89} & 61.24 & 76.55 & -18.78 \\
O-LoRA~\cite{olora2023} & EMNLP 2023
& 77.50 & 78.07 & 44.50 & 63.13 & 64.73 & \underline{58.16}
& \underline{76.01} & 64.35 & 78.02 & -13.99 \\
MoELoRA~\cite{coin2024} & NeurIPS 2024
& 70.07 & 77.70 & 44.69 & 50.03 & 54.03 & 57.34
& 71.17 & 58.98 & 75.08 & -14.63 \\
ModalPrompt~\cite{modalprompt2025} & EMNLP 2025
& 51.07 & 87.27 & 48.11 & 39.23 & 46.57 & 42.93
& 52.65 & 52.53 & 57.67 & \textbf{-0.15} \\
CL-MoE~\cite{clmoe2025} & CVPR 2025
& 66.33 & 77.00 & 44.78 & 51.87 & 53.53 & 57.42
& 71.46 & 58.49 & 74.19 & -15.56 \\
HiDe~\cite{hide2025} & ACL 2025
& 84.03 & 90.73 & 44.43 & 58.93 & 41.37 & 54.25
& 69.96 & 62.29 & 77.32 & -9.20 \\
SEFE~\cite{sefe2025} & ICML 2025
& 80.83 & 78.00 & 47.01 & \underline{69.63} & 65.83 & 57.92
& 75.98 & 66.54 & 78.76 & -11.33 \\
DISCO~\cite{federated2025} & ICCV 2025
& 87.43 & \underline{93.07} & 46.96 & 68.13 & 65.70 & 56.69
& 75.87 & 69.66 & \underline{81.60} & -7.45 \\
KeepLoRA~\cite{keeplora2026} & ICLR 2026
& 82.43 & 86.70 & 46.54 & 67.80 & 66.40 & 57.18
& -- & 67.84 & -- & -- \\
Octopus~\cite{liu2026octopus} & CVPR 2026
& \underline{88.20} & 93.03 & \textbf{58.46} & 60.50
& \underline{69.00} & 56.87
& 70.60 & \underline{71.01} & -- & -- \\
SAME~\cite{same2026} & ICML 2026
& 83.83 & 91.40 & 51.33 & 65.27 & 53.50 & 57.43
& -- & 67.12 & -- & -- \\
SiGMA~\cite{sigma2026} & ECCV 2026
& 83.40 & \textbf{94.07} & 54.83 & 51.03 & 57.83 & 55.87
& -- & 66.17 & -- & -- \\
DRAPE~\cite{drape2026} & NeurIPS 2026
& 85.07 & 91.60 & \underline{55.68} & 62.53 & 65.77 & 55.82
& -- & 69.41 & -- & -- \\
\hline
\textbf{\method{}} & --
& \textbf{91.33} & 92.60 & 54.40 & \textbf{74.63}
& \textbf{76.10} & 56.84
& 75.54 & \textbf{74.32} & \textbf{83.01} & \underline{-1.47} \\
\bottomrule
\end{tabular}
\label{tab:ucit-main}
\end{table*}
\begin{table*}[!t]
\caption{Comparison with existing methods on TriGap.
The best results are in bold, and the second-best results are underlined.}
\label{tab:trigap-main}
\centering
\footnotesize
\setlength{\tabcolsep}{1.4pt}
\renewcommand{\arraystretch}{1.12}
\begin{tabular}{lccccccccccc|>{\columncolor{gray!10}}c}
\toprule
Method & Venue & PMCVQA & DocVQA &  ChartQA &IconQA & InfographicVQA & ArxivQA & Roadside & ChemVQA & FloodNetVQA & CLEVR & Avg. \\
\hline

Zero-shot & -- 
& 35.40 
& 12.68 
& 9.36 
& 19.27 
& 5.06 
& 53.77 
& 7.40 
& 5.30 
& 47.41 
& 20.37 
& 21.60 \\

FT-LoRA~\cite{lora2022} & ICLR 2022 
& 34.20 
& 23.32 
& 9.84 
& 37.07 
& 23.53 
& 83.83 
& 7.00 
& 12.70 
& 80.31 
& 60.27 
& 37.21 \\

MoELoRA~\cite{coin2024} & NeurIPS 2024 
& 39.03 
& 37.49 
& 12.44 
& 43.43 
& 35.17 
& 90.90 
& 7.93 
& 20.70 
& \textbf{90.41} 
& \underline{67.00} 
& 44.45 \\

HiDe~\cite{hide2025} & ACL 2025 
& 37.00 
& 33.20 
& 10.52 
& 41.97 
& 24.09 
& 79.20 
& 7.73 
& 11.17 
& 57.39 
& 23.00 
& 32.53 \\

ModalPrompt~\cite{modalprompt2025} & EMNLP 2025 
& 38.23 
& 38.23 
& 11.92 
& 44.73 
& 37.37 
& 84.47 
& 10.13 
& 12.43 
& 71.52 
& 52.50 
& 40.15 \\

CL-MoE~\cite{clmoe2025} & CVPR 2025 
& 40.53 
& 36.79 
& 13.72 
& 52.70 
& 32.27 
& \underline{93.00} 
& 7.77 
& 18.33 
& 80.09 
& 65.90 
& 44.11 \\

DISCO~\cite{federated2025} & ICCV 2025 
& \underline{42.03} 
& 43.50 
& \textbf{18.01} 
& 63.13 
& 38.23 
& 91.27 
& \underline{11.02} 
& \underline{22.13} 
& 80.25 
& 55.87 
& \underline{46.54} \\

SAME~\cite{same2026} & ICML 2026 
& 41.60 
& \underline{43.87} 
& \underline{17.56} 
& \underline{64.03} 
& \underline{39.57} 
& 90.46 
& 10.83 
& 21.77 
& 81.09 
& 54.50 
& 46.53 \\

\hline
\textbf{\method{}} & -- 
& \textbf{44.23} 
& \textbf{50.00} 
& 17.52 
& \textbf{68.10} 
& \textbf{42.25} 
& \textbf{93.93} 
& \textbf{12.67} 
& \textbf{33.50} 
& \underline{82.82} 
& \textbf{68.33} 
& \textbf{51.33} \\

\bottomrule
\vspace{-6mm}
\end{tabular}
\end{table*}

\section{Experiments}
\label{sec:experiments}

\subsection{Experimental Setups}
\label{sec:experimental-setup}

\subsubsection{Benchmarks}
We evaluate our method on four publicly available multimodal
continual instruction-tuning benchmarks.
(a) UCIT~\cite{hide2025,mcitlib2025} contains ImageNet-R, ArXivQA,
VizWiz, IconQA, CLEVR-Math, and Flickr30k. It combines recognition,
question answering, mathematical reasoning, and captioning, including
related instructions associated with different tasks.
(b) TriGap~\cite{same2026} presents ten tasks: PMCVQA, DocVQA,
ChartQA, IconQA, InfographicVQA, ArXivQA, Roadside, ChemVQA,
FloodNetVQA, and CLEVR. Its longer, specialized task stream evaluates
adaptation across heterogeneous domains.
(c) CVIT-Single~\cite{smolora2025}, abbreviated as CVIT, contains
ScienceQA, TextVQA, Flickr30k, ImageNet, GQA, and VQAv2, with one
instruction type per task.
(d) CoIN~\cite{coin2024} contains ScienceQA, TextVQA, ImageNet, GQA,
VizWiz, referring expression comprehension (REC), VQAv2, and OCR-VQA.
It combines visual recognition, grounding, and question answering.
We use the prescribed task orders and train/test splits. Training accesses
only the arriving task, and evaluation covers all learned tasks without
their task identifiers.

\subsubsection{Implementation Details}
The main backbone is LLaVA-v1.5-7B~\cite{liu2024llava15}, with
InternVL-Chat-7B~\cite{internvl2024} additionally evaluated on UCIT.
The visual encoder and language-model backbone are frozen, and only
the selected root's projector and LoRA factors are optimized.
A frozen CLIP text encoder~\cite{radford2021clip} constructs instruction
prototypes. For LLaVA, we use AdamW and one epoch per task, with
learning rates of $2\times10^{-4}$ for LoRA and $2\times10^{-5}$ for
the projector. The default expert rank is $16$. The per-device
batch size is $32$ with two gradient-accumulation steps on one A800 GPU.
The semantic threshold is $\Theta=0.95$, the amplitude exponent is
$\alpha=2$, and the numerical floor is $\varepsilon=10^{-12}$.
The calibration coefficient $\kappa_t$ is estimated from participating
module ratios at each synthesis event. 

\subsubsection{Evaluation Metrics}
Let $a_{t,j}$ be the score on task $j$ after learning task $t$.
\begin{equation}
\begin{aligned}
\mathrm{MFT}&=\frac1T\sum_{t=1}^{T}a_{t,t},&
\mathrm{MFN}&=\frac1T\sum_{j=1}^{T}a_{T,j},\\
\mathrm{MAA}&=\frac1T\sum_{t=1}^{T}\frac1t\sum_{j=1}^{t}a_{t,j},&
\mathrm{BWT}&=\frac1{T-1}\sum_{j=1}^{T-1}(a_{T,j}-a_{j,j}).
\end{aligned}
\label{eq:exp-metrics}
\end{equation}
MFT measures acquisition immediately after learning each task. MFN
measures the final retained task performance, and MAA averages performance
throughout the stream. BWT measures the change in earlier-task scores
relative to their acquisition stages; higher values indicate better
retention. Jointly considering these quantities distinguishes effective
acquisition and retention from limited learning with little subsequent
change. For the shared-root pair on UCIT, we additionally report
$\mathrm{Sim\mbox{-}Acc}
=(a_{T,\mathrm{VizWiz}}+a_{T,\mathrm{Flickr30k}})/2$.
This measures how well the shared expert serves both tasks. For CVIT, AP and MAP denote final and stage-averaged performance under
the benchmark's native task scores, including CIDEr for
Flickr30k~\cite{smolora2025}. TriGap and CoIN report final task-wise
scores and their macro-average. ScienceQA uses overall test accuracy.
We compare aggregates within each benchmark because their task-score
conventions differ. For class-incremental learning, $A_{\mathrm{Last}}$ denotes the accuracy on all learned classes after the final task, while
$A_{\mathrm{Avg}}$ averages the accuracy on the classes learned
so far across all training stages.

\begin{table*}[!t]
\caption{Comparison with existing methods on CVIT-Single.
The best results are in bold, and the second-best results are underlined.}
\label{tab:cvit-main}
\centering
\scriptsize
\setlength{\tabcolsep}{8pt}
\renewcommand{\arraystretch}{1.08}
\small
\begin{tabular}{lccccccc|>{\columncolor{gray!10}}c>{\columncolor{gray!10}}cc}
\toprule
Method & Venue & Science & Text & Flickr & ImageNet & GQA & VQAv2 & AP & MAP & BWT \\
\hline

Zero-shot & -- 
& 52.72 
& 2.95 
& 52.64 
& 22.10 
& 2.73 
& 0.65 
& 22.30 
& -- 
& -- \\

SeqLoRA~\cite{lora2022} & ICLR 2022 
& 55.31 
& 50.22 
& 33.89 
& 22.73 
& 50.52 
& 64.61 
& 46.21 
& 57.41 
& -48.10 \\

MoELoRA~\cite{coin2024} & NeurIPS 2024 
& 55.01 
& 48.87 
& 32.04 
& 22.00 
& 50.03 
& 63.64 
& 45.27 
& 56.16 
& -48.05 \\

DoRA~\cite{dora2024} & ICML 2024 
& 51.26 
& 46.36 
& 36.41 
& 28.24 
& 45.29 
& 56.87 
& 44.07 
& 65.03 
& -31.12 \\

C-LoRA~\cite{clora2023} & TMLR 2024 
& 57.25 
& 38.70 
& 56.50 
& 25.27 
& 42.89 
& 54.06 
& 45.78 
& 57.04 
& -19.58 \\

EWC~\cite{kirkpatrick2017} & PNAS 2017 
& 57.04 
& 50.02 
& 32.96 
& 22.85 
& 50.16 
& 64.54 
& 46.26 
& 56.19 
& -49.71 \\

Eproj~\cite{he2026continual} & TIP 2026 
& 65.29 
& 52.87 
& 148.19 
& 39.45 
& 28.06 
& 57.86 
& 65.29 
& 73.53 
& -14.02 \\

SMoLoRA~\cite{smolora2025} & ICCV 2025 
& 77.36 
& 58.29 
& 151.99 
& 95.35 
& 51.96 
& \textbf{65.71} 
& 83.44 
& {84.85} 
& {-3.23} \\

LiLoRA~\cite{che2026lora} & AAAI 2026 
& \underline{77.88} 
& \underline{58.83} 
& \underline{152.93}
& \underline{96.02}
& \underline{58.28}
& \underline{65.33} 
& \underline{84.88}
& \underline{87.70} 
& \underline{-3.13} \\

\hline
\textbf{\method{}} & -- 
& \textbf{88.14} 
& \textbf{64.22} 
& \textbf{176.44} 
& \textbf{97.37} 
& \textbf{59.75} 
& 64.31 
& \textbf{91.71} 
& \textbf{95.08} 
& \textbf{-0.74} \\

\bottomrule
\end{tabular}
\end{table*}
\begin{table*}[!t]
\caption{Comparison with existing methods on CoIN. 
The best results are in bold, and the second-best results are underlined.}
\label{tab:coin-main}
\centering
\scriptsize
\setlength{\tabcolsep}{4pt}
\renewcommand{\arraystretch}{1.08}
\small
\begin{lrbox}{\additionaltablebox}
\begin{tabular}{lccccccccc|>{\columncolor{gray!10}}c}
\toprule
Method & Venue & ScienceQA & TextVQA & ImageNet & GQA & VizWiz & REC & VQAv2 & OCR-VQA & Accuracy \\
\hline
MoELoRA~\cite{coin2024} & NeurIPS 2024
& 62.02 & 52.05 & 37.21 & 53.12 & 43.32 & 33.22 & 57.92 & 65.75
& 50.58 \\
Continual LLaVA~\cite{cao2024continualllava} & --
& 58.67 & 49.99 & 57.66 & \underline{62.53} & 42.32 & 16.25
& 64.33 & \textbf{74.91} & 53.33 \\
ModalPrompt~\cite{modalprompt2025} & EMNLP 2025
& 68.42 & 56.40 & 41.13 & 61.11 & 50.13 & 36.69 & 66.90 & 59.68
& 55.06 \\
SEFE~\cite{sefe2025} & ICML 2025
& 75.35 & 58.66 & 83.10 & 54.25 & 48.85 & 16.75 & 65.35 & 66.25
& 58.57 \\
ProgLoRA~\cite{progressivelora2025} & ACL 2025
& 74.84 & 51.83 & 83.90 & 49.93 & 53.87 & 31.19 & 62.71 & 64.44
& 59.09 \\
HiDe-LLaVA~\cite{hide2025} & ACL 2025
& 73.20 & 56.92 & 69.28 & 61.33 & 50.76 & 59.18
& \underline{67.12} & 64.76 & 62.82 \\
PCLR~\cite{pclr2026} & ICLR 2026
& 78.33 & 58.24 & 86.08 & 58.14 & 57.61 & 33.04 & 64.17 & 61.92
& 62.19 \\
Octopus~\cite{liu2026octopus} & CVPR 2026
& 79.72 & 53.60 & 82.18 & 60.50 & 51.49 & 62.17 & 66.38 & 64.72
& 65.10 \\
$\phi$-DPO$$~\cite{phidpo2026} & CVPR 2026
& 77.84 & 59.17 & 95.61 & \textbf{64.32} & 54.55 & 60.74
& \textbf{69.99} & \underline{68.69} & \underline{68.86} \\
SAME~\cite{same2026} & ICML 2026
& 78.35 & \underline{60.69} & 90.21 & 61.70 & 54.13 & 59.87
& 66.04 & 63.59 & 66.82 \\
Hyper-LLaVA~\cite{hyperllava2026} & ICML 2026
& 77.10 & 59.02 & 93.49 & 60.54 & \underline{57.88} & 51.81
& 66.76 & 65.13 & 66.47 \\
MoBLoRA~\cite{moblora2026} & ACL 2026
& \underline{84.96} & 60.46 & \textbf{96.87} & 60.07
& \textbf{61.70} & 32.95 & 64.82 & 60.85 & 65.34 \\
DRAPE~\cite{drape2026} & NeurIPS 2026
& 70.67 & 59.61 & 94.16 & 61.37 & 54.43 & \underline{67.92}
& 66.53 & 65.11 & 67.48 \\
\hline
\textbf{\method{}} & --
& \textbf{87.29} & \textbf{64.03} & \underline{96.53} & 58.33
& 50.96 & \textbf{76.10} & 64.88 & 64.07 & \textbf{70.27} \\
\bottomrule
\end{tabular}
\end{lrbox}
\ifdim\wd\additionaltablebox>\textwidth
\resizebox{\textwidth}{!}{\usebox{\additionaltablebox}}
\else\usebox{\additionaltablebox}\fi
\end{table*}

\subsection{Main Results}
\label{sec:main-results}

We compare with three groups of continual adaptation methods.
\emph{Shared and constrained updates} include sequential LoRA
fine-tuning~\cite{lora2022}, subspace-constrained O-LoRA~\cite{olora2023},
magnitude--direction adaptation with DoRA~\cite{dora2024},
regularized C-LoRA and EWC~\cite{clora2023,kirkpatrick2017}, and
history-free gradient orthogonalization with Octopus~\cite{liu2026octopus}. KeepLoRA~\cite{keeplora2026} restricts adaptation to a residual
subspace that protects pretrained and historical knowledge, while
SiGMA~\cite{sigma2026} uses parameter signs to guide adaptation and merging.
These methods primarily regulate changes to adaptation parameters.
\emph{Prompt and increment-based methods} include modality-guided
ModalPrompt~\cite{modalprompt2025}, input-conditioned cross-modal
prompt generation with DRAPE~\cite{drape2026}, intrinsic and contextual
increment retrieval in Continual LLaVA~\cite{cao2024continualllava}, and
projector expansion in EProj~\cite{he2026continual}.
\emph{Expert-based and structured methods} include
MoELoRA~\cite{coin2024}, dual-momentum CL-MoE~\cite{clmoe2025},
hierarchical HiDe-LLaVA~\cite{hide2025}, modular DISCO~\cite{federated2025},
and progressive ProgLoRA~\cite{progressivelora2025}. This group also
includes SMoLoRA~\cite{smolora2025}, which separates visual and instruction
adaptation; LiLoRA~\cite{che2026lora}, which factorizes shared and
task-specific updates; SEFE~\cite{sefe2025}, which addresses response-style
and knowledge forgetting; and SAME~\cite{same2026}, which controls
stability across the instruction stream. PCLR~\cite{pclr2026} progressively compresses and integrates knowledge within a LoRA rank pool, while MoBLoRA~\cite{moblora2026} reconstructs task-specific adaptations from shared orthogonal bases. Hyper-LLaVA~\cite{hyperllava2026} combines hyperbolic task-distribution matching with uncertainty-aware modality weighting. 
We additionally compare with $\phi$-DPO~\cite{phidpo2026}, which uses
pairwise preference supervision and an imbalance-aware objective.
Zero-shot and Individual tuning are untuned and task-isolated references,
respectively, where reported.

Table~\ref{tab:ucit-main} shows that \method{} achieves the best
MFN and MAA on UCIT, improving MFN by 3.31 points over the
strongest competing result. While LoRA-FT exhibits substantial
forgetting despite strong immediate acquisition, and ModalPrompt's
small backward change accompanies weaker task performance, our
method maintains a more favorable acquisition--retention trade-off.
This improvement is achieved while VizWiz and Flickr30k
share one executable root. Table~\ref{tab:trigap-main} reports a
4.79-point improvement in the final average score on TriGap.
Its relatively well-separated task profiles
(Fig.~\ref{fig:motivation}) suggest limited instruction-based
task ambiguity. Accordingly, the tasks retain distinct roots,
and ARS is not invoked; this setting primarily
examines task-specific acquisition and soft inference.
On CVIT, Table~\ref{tab:cvit-main} shows the best AP, MAP,
and BWT, with AP exceeding the strongest competing result by
6.83 points. The improvements over
SMoLoRA~\cite{smolora2025} and LiLoRA~\cite{che2026lora}
are accompanied by stronger retention of earlier scientific
question answering and captioning capabilities.
Table~\ref{tab:coin-main} further shows the highest final average
on CoIN, exceeding DRAPE and $\phi$-DPO by 2.79 and 1.41 points,
respectively. The improvement is accompanied by strong scientific
question answering, text understanding, and grounding performance,
indicating effective retention across different instruction requirements.
Together, these results demonstrate the effectiveness of the
framework across task streams requiring either shared or separate
experts, while the following ablations examine the contributions
of expert organization and ARS.

\subsection{Ablation Studies}
\label{sec:ablations}

We conduct ablation studies to examine how root-level expert
organization and ARS contribute to continual instruction
learning. We first progressively introduce expert organization, amplitude
share, and change confidence, and then compare alternative synthesis
rules. Further analyses investigate root-level execution, empirical
confidence calibration, fixed-rank restoration, and projector adaptation.

\subsubsection{Component Ablation}
\providecolor{sharedtask}{RGB}{0,100,60}
\providecolor{pairedscore}{RGB}{128,70,170}
\begin{table*}[!t]
\centering
\caption{Progressive component ablation on UCIT.
Sem. Expert denotes semantic expert organization, including expert
allocation and root-level inference; Amp. and Conf. denote amplitude
share and change confidence. Sim-Acc averages the two starred tasks. 
The all-cross row corresponds to sequential LoRA fine-tuning.
Best and second-best results are boldfaced and underlined.}
\label{tab:component-ablation}
\begingroup
\small
\setlength{\tabcolsep}{2pt}
\renewcommand{\arraystretch}{1.08}
\begin{tabular}{@{}ccc cc>{\color{sharedtask}}c cc>{\color{sharedtask}}c
c >{\columncolor{gray!10}}c >{\columncolor{gray!10}}c
c >{\color{pairedscore}}c@{}}
\toprule
\shortstack{Sem. Expert} & Amp. & Conf.
& ImageNet-R & ArXivQA & \multicolumn{1}{r}{VizWiz$^{\ast}$} & IconQA
& CLEVR-Math & \multicolumn{1}{r}{Flickr30k$^{\ast}$}
& MFT$\uparrow$ & MFN$\uparrow$ & MAA$\uparrow$
& BWT$\uparrow$ & \multicolumn{1}{r}{Sim-Acc$\uparrow$} \\
\midrule
\textbf{\ding{55}} & \textbf{\ding{55}} & \textbf{\ding{55}}
& 58.03 & 77.63 & 44.39 & 67.40 & 61.77 & \textbf{58.22}
& \textbf{76.89} & 61.24 & 76.55 & $-18.78$ & 51.31 \\
\textbf{\ding{51}} & \textbf{\ding{55}} & \textbf{\ding{55}}
& 91.23 & \textbf{93.07} & 48.00 & \underline{75.00} & 75.40 & 56.18
& 75.29 & 73.15 & 82.81 & $-2.57$ & 52.09 \\
\textbf{\ding{51}} & \textbf{\ding{51}} & \textbf{\ding{55}}
& \textbf{91.47} & 92.37 & \underline{51.67} & \textbf{75.27}
& \underline{75.97} & \underline{57.43}
& 75.43 & \underline{74.03} & \underline{82.90}
& $\underline{-1.68}$ & \underline{54.55} \\
\textbf{\ding{51}} & \textbf{\ding{51}} & \textbf{\ding{51}}
& \underline{91.33} & \underline{92.60} & \textbf{54.40} & 74.63
& \textbf{76.10} & 56.84
& \underline{75.54} & \textbf{74.32} & \textbf{83.01}
& $\mathbf{-1.47}$ & \textbf{55.62} \\
\bottomrule
\end{tabular}
\endgroup
\end{table*}
\providecolor{sharedtask}{RGB}{0,100,60}
\providecolor{pairedscore}{RGB}{128,70,170}
\begin{table*}[!t]
\centering
\caption{Comparison of synthesis rules on UCIT.
The six task columns report final scores; Sim-Acc averages the two starred tasks.
The best and second-best results are boldfaced and underlined, respectively.}
\label{tab:synthesis-analysis}
\begingroup
\small
\setlength{\tabcolsep}{2.6pt}
\renewcommand{\arraystretch}{1.02}
\begin{tabular}{@{}l cc>{\color{sharedtask}}c cc>{\color{sharedtask}}c
c >{\columncolor{gray!10}}c >{\columncolor{gray!10}}c
c >{\color{pairedscore}}c@{}}
\toprule
Rule & ImageNet-R & ArXivQA & \multicolumn{1}{r}{VizWiz$^{\ast}$} & IconQA
& CLEVR-Math & \multicolumn{1}{c}{Flickr30k$^{\ast}$}
& MFT$\uparrow$ & MFN$\uparrow$ & MAA$\uparrow$
& BWT$\uparrow$ & \multicolumn{1}{r}{Sim-Acc$\uparrow$} \\
\midrule
Direct update
& \textbf{91.53} & \textbf{93.00} & 46.11 & \textbf{75.87}
& 76.20 & \textbf{57.99}
& \textbf{75.92} & 73.45 & \textbf{83.07} & $-2.97$ & 52.05 \\
Fisher weighting
& \underline{91.50} & 92.67 & 47.79 & 73.30
& \underline{76.57} & 55.82
& 75.21 & 72.94 & 82.73 & $-2.72$ & 51.81 \\
Uniform interpolation
& 91.43 & 92.23 & 51.28 & \underline{75.70}
& 76.27 & 56.81
& 75.54 & 73.95 & 82.91 & $-1.90$ & 54.05 \\
Utility
& 91.47 & \underline{92.70} & \underline{51.45} & 75.60
& \textbf{76.73} & \underline{57.20}
& \underline{75.76} & \underline{74.19} & \textbf{83.07}
& $\underline{-1.88}$ & \underline{54.33} \\
\textbf{ARS (ours)}
& 91.33 & 92.60 & \textbf{54.40} & 74.63 & 76.10 & 56.84
& 75.54 & \textbf{74.32} & \underline{83.01}
& $\mathbf{-1.47}$ & \textbf{55.62} \\
\bottomrule
\end{tabular}
\endgroup
\end{table*}

Table~\ref{tab:component-ablation} starts from sequential LoRA
fine-tuning, which continually updates a single adaptation state on
arriving tasks. This baseline achieves strong immediate acquisition,
but its substantially lower final performance indicates that learning
the current task does not ensure preservation of earlier capabilities.
The Sem. Expert configuration instead maintains instruction-conditioned
projector--LoRA experts and restricts training to the selected expert,
with task probabilities aggregated by root at inference.
This limits the exposure of unrelated task functions to subsequent
updates. The improvement in MFN from 61.24 to 73.15, together with the
stronger final recognition and question-answering scores, demonstrates
the benefit of semantically organized adaptation states.

Amplitude-based synthesis further combines the stored and newly trained
effective LoRA weights within a reused expert. 
Compared with expert
organization alone, this configuration improves the final performance
of both VizWiz and Flickr30k. This suggests that organizing experts
and consolidating their parameters address different requirements:
the former determines which adaptation state serves a task, whereas
the latter controls how that state incorporates subsequent learning. Change confidence additionally regulates the amplitude-proposed update
according to its strength relative to the module reference. Stronger
relative changes retain a larger proportion of the proposal, while
weaker changes receive greater contraction. Adding this regulation
improves Sim-Acc by 1.07 points and yields better final average
performance and backward transfer. 
This trade-off is consistent with the purpose of controlling adaptation
within a shared expert: 
incorporating new instruction requirements while reducing
the disruption to previously acquired capabilities caused
by new-task learning.
The complete configuration consequently provides the strongest final
and shared-task performance among the component variants.

\begin{table}[!t]
\centering
\caption{CIL no-replay comparison on 10S-CUB and 10S-ImageNet-A with ViT B/16 backbone. Results are reported as mean $\pm$ standard deviation. Best and second-best results are boldfaced and underlined.}
\label{tab:cil-ablation}
\begingroup
\small
\setlength{\tabcolsep}{3pt}
\renewcommand{\arraystretch}{1.08}
\resizebox{\columnwidth}{!}{
\begin{tabular}{@{}lcccc@{}}
\toprule
& \multicolumn{2}{c}{10S-CUB}
& \multicolumn{2}{c}{10S-ImageNet-A} \\
\cmidrule(lr){2-3}
\cmidrule(l){4-5}
Method
& $A_{\mathrm{Last}}\uparrow$
& $A_{\mathrm{Avg}}\uparrow$
& $A_{\mathrm{Last}}\uparrow$
& $A_{\mathrm{Avg}}\uparrow$ \\
\midrule
Seq-FT
& $64.82\pm2.30$
& $75.28\pm1.56$
& $52.56\pm2.08$
& $63.36\pm1.32$ \\

Amplitude-only
& $\underline{79.37\pm0.64}$
& $\underline{87.50\pm0.74}$
& $\underline{57.13\pm1.27}$
& $\underline{65.80\pm1.81}$ \\

\textbf{ARS}
& $\mathbf{80.52\pm1.24}$
& $\mathbf{88.49\pm0.73}$
& $\mathbf{58.22\pm1.13}$
& $\mathbf{66.51\pm1.83}$ \\
\bottomrule
\end{tabular}
}
\endgroup
\end{table}

Table~\ref{tab:cil-ablation} further examines the application of
ARS to class-incremental learning on ten-task CUB and ImageNet-A.
Sequential LoRA fine-tuning directly retains the newly trained
state, whereas amplitude-only synthesis combines the stored and
acquired states according to their relative weight magnitudes.
ARS additionally uses change confidence to regulate the retention
of the proposed changes.
Among the reported configurations, ARS achieves the highest final
and stage-averaged accuracy on both datasets.
Its strong performance throughout the task sequence, together
with the retained accuracy after the final task, demonstrates
its applicability to successive classification tasks beyond
multimodal instruction tuning.

\subsubsection{Alternative Synthesis Rules}

Table~\ref{tab:synthesis-analysis} compares alternative consolidation
rules for LoRA experts. Direct updating keeps the acquired state,
uniform interpolation assigns equal weights, and Fisher weighting
combines independently acquired LoRA factors using diagonal
gradient-derived importance~\cite{matena2022fisher,kirkpatrick2017}.
Utility instead weights each connection by its absolute weight
and mean absolute input activation~\cite{dohare2024plasticity}. 
Direct updating favors immediate acquisition but
degrades earlier VizWiz performance. Utility improves the shared-task
balance over direct updating and Fisher weighting, while ARS further
improves Sim-Acc from $54.33$ to $55.62$. 
The confidence-free amplitude variant in
Table~\ref{tab:component-ablation} also obtains a comparable Sim-Acc of
$54.55$ without storing activations, supporting the weight-based
importance construction.

\begin{figure*}[!t]
\centering
\includegraphics[width=\textwidth]{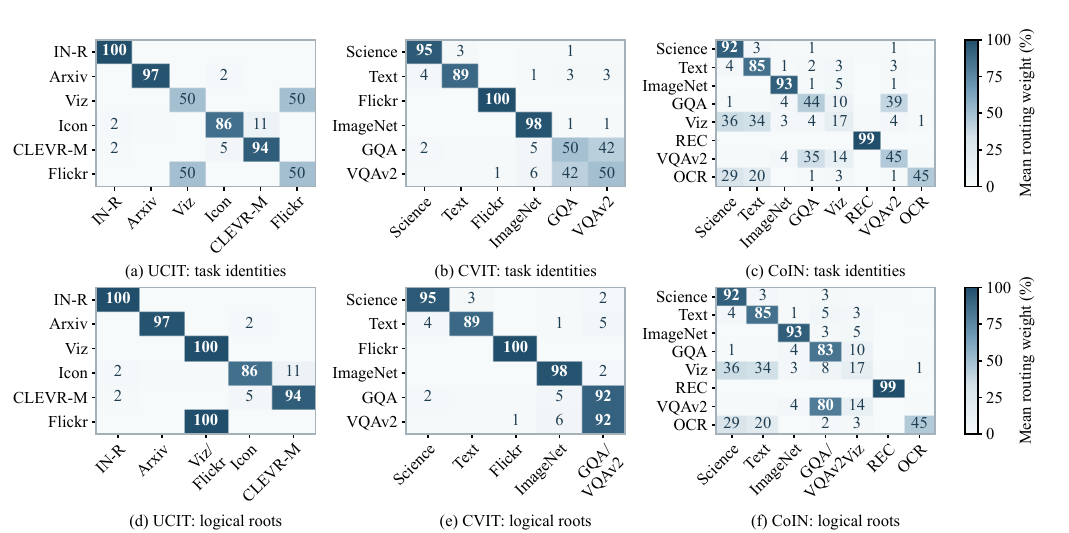}
\caption{Mean soft routing weights over task identities (top) and logical
roots (bottom) on UCIT, CVIT, and CoIN.
Rows denote true tasks. Root weights aggregate the probabilities of
identities assigned to the same root.}
\label{fig:root_routing}
\end{figure*}

\providecolor{pairedscore}{RGB}{128,70,170}
\begin{table}[!t]
\centering
\caption{Importance-estimation cost and shared-task performance on UCIT.
$N_{\rm stat}$ and storage assume one FP32 importance summary per task.
$G_{\rm data}$ measures additional forward and backward GFLOPs per
640-token example. Sim-Acc is the mean final score of VizWiz and Flickr30k.}
\label{tab:importance-cost}
\begingroup
\small
\setlength{\tabcolsep}{3pt}
\renewcommand{\arraystretch}{1.05}
\begin{tabular}{@{}l ccc >{\color{pairedscore}}c@{}}
\toprule
Importance rule
& $N_{\rm stat}\downarrow$
& Storage (MiB)$\downarrow$
& $G_{\rm data}\downarrow$
& Sim-Acc$\uparrow$ \\
\midrule
Fisher weighting
& 239\,861\,760 & 915.00 & 17\,434.28 & 51.81 \\
Utility
& \underline{6\,832\,128} & \underline{26.06}
& \underline{8\,937.79} & \underline{54.33} \\
\textbf{ARS (ours)}
& \textbf{0} & \textbf{0.00}
& \textbf{0.00} & \textbf{55.62} \\
\bottomrule
\end{tabular}
\endgroup
\end{table}

Table~\ref{tab:importance-cost} compares the storage and computation
required for importance estimation. Storing one FP32 summary for each
of the six tasks requires $26.06$ MiB for Utility and $915.00$ MiB for
Fisher weighting, whereas ARS retains no additional activation or
gradient summaries. Utility estimates input activations through an
additional forward pass, and Fisher weighting also computes gradients.
The table reports their additional data-dependent computation per
640-token example over the supported operators.
ARS computes importance from the available weights; its squared-amplitude
calculation costs $25.90$ GFLOPs for the two weight states without an
additional data pass. Together with its higher Sim-Acc, this provides a
favorable balance between performance, storage, and computation.
\subsubsection{Identity Ambiguity and Root-Level Execution}

Fig.~\ref{fig:root_routing} shows how task probabilities are combined
for shared experts. On UCIT, VizWiz and Flickr30k receive separate
task probabilities, and these probabilities are added to weight the
expert they share. GQA and VQAv2 show the corresponding behavior on
CVIT and CoIN. Redistributing probability between tasks that share an
expert therefore leaves its contribution unchanged when their sum is
fixed. 

\subsubsection{Effect of Change Confidence}

\begin{figure*}[!t]
\centering
\includegraphics[width=\textwidth]{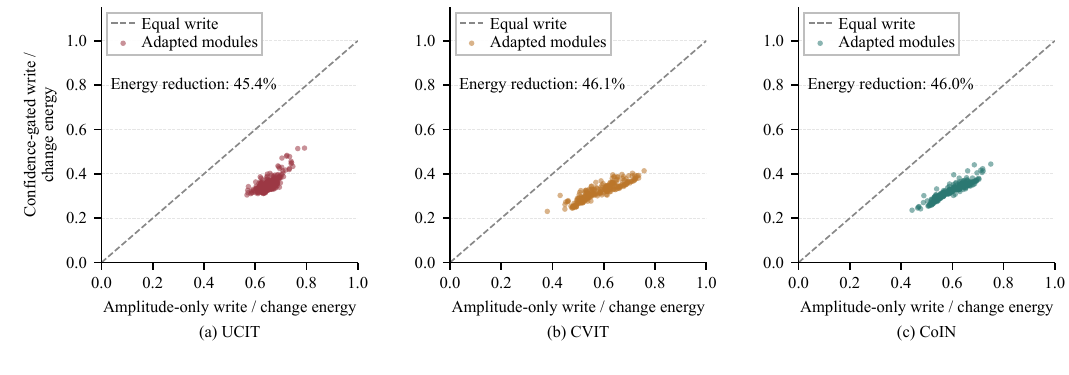}
\vspace{-8mm}
\caption{Confidence-gated write contraction on UCIT, CVIT, and CoIN.
Each point represents an adapted module. Proposed and gated write energies
are normalized by acquisition-change energy; the dashed line denotes
equal write energy.}
\label{fig:contraction}
\end{figure*}

Fig.~\ref{fig:contraction} compares the amplitude-only proposal
$\boldsymbol{\Lambda}^{\rm amp}\odot\mathbf{D}$ with the gated write
$\boldsymbol{\rho}\odot\boldsymbol{\Lambda}^{\rm amp}\odot\mathbf{D}$.
Each point is a module: the horizontal and vertical coordinates are the
squared Frobenius norms of these two writes, respectively, divided by
$\|\mathbf{D}\|_F^2$. The points lie below the equal-write line,
illustrating the contraction in Eq.~(\ref{eq:post-rank-bound}).
The different degrees of contraction show that the gate responds to
module-specific changes.
Together with the ablation results in
Table~\ref{tab:component-ablation}, these observations indicate that
change-dependent update retention improves the final performance of
the shared expert, providing a better balance between new-task
adaptation and preservation of previously learned capabilities.

\begin{table}[!t]
\caption{Share of squared acquisition-change energy carried by the
largest 1\%, 5\%, 10\%, and 20\% of entries on UCIT, CVIT, and CoIN.}
\label{tab:energy-concentration}
\centering
\scriptsize
\setlength{\tabcolsep}{9.5pt}
\renewcommand{\arraystretch}{1.08}
\small
\begin{lrbox}{\additionaltablebox}
\begin{tabular}{c|cccc}
\toprule
Benchmark 
& Top 1\% 
& Top 5\% 
& Top 10\% 
& Top 20\% \\
\midrule

UCIT
& 12.38\% 
& 34.52\% 
& 50.68\% 
& 70.33\%  \\

CVIT
& 12.31\% 
& 34.67\% 
& 50.98\% 
& 70.68\%  \\

CoIN
& 11.79\% 
& 33.62\% 
& 49.80\% 
& 69.65\%  \\

\bottomrule
\end{tabular}
\end{lrbox}
\ifdim\wd\additionaltablebox>\textwidth
\usebox{\additionaltablebox}
\else
\usebox{\additionaltablebox}
\fi
\end{table}

Table~\ref{tab:energy-concentration} measures the energy carried by
the largest squared changes. Across the three benchmarks, relatively
few entries carry a disproportionate share of acquisition-change energy.
This supports entry-wise retention and a module-relative reference:
equal treatment of all entries would ignore the uneven change structure.
These results support regulating amplitude-based updates according to
relative change strength. The confidence ablation in
Table~\ref{tab:component-ablation} further demonstrates the benefit
of this regulation for final task performance.

\definecolor{deepgreen}{RGB}{0,100,60}
\definecolor{deeppurple}{RGB}{128,70,170}

\begin{table*}[!t]
\centering
\caption{Confidence-calibration sensitivity on UCIT.
The six task columns report final performance.
Sim-Acc averages the final scores of VizWiz and Flickr30k.
The best results are boldfaced, and the second-best results are underlined.}
\label{tab:kappa-sensitivity}

\scriptsize
\setlength{\tabcolsep}{5pt}
\renewcommand{\arraystretch}{1}
\small

\begin{tabular*}{\textwidth}{
@{\extracolsep{\fill}}
l| c c c c c c |
c
>{\columncolor{gray!10}}c
>{\columncolor{gray!10}}c c c @{}
}
\toprule
Setting
& ImageNet-R
& ArXivQA
& VizWiz$^{\ast}$
& IconQA
& CLEVR-Math
& Flickr30k$^{\ast}$
& MFT$\uparrow$
& MFN$\uparrow$
& MAA$\uparrow$
& BWT$\uparrow$
& Sim-Acc$\uparrow$ \\
\midrule

$\kappa=0$ (w/o $\rho$)
& 91.47
& 92.37
& \textcolor{deepgreen}{51.67}
& \textbf{75.27}
& \underline{75.97}
& \textcolor{deepgreen}{\textbf{57.43}}
& \underline{75.43}
& 74.03
& 82.90
& $-1.68$
& \textcolor{deeppurple}{54.55} \\

$\kappa=0.375$
& 91.33
& \underline{92.60}
& \textcolor{deepgreen}{\textbf{54.40}}
& \underline{74.63}
& \textbf{76.10}
& \textcolor{deepgreen}{56.84}
& \textbf{75.54}
& \textbf{74.32}
& \underline{83.01}
& $-1.47$
& \textcolor{deeppurple}{\textbf{55.62}} \\

$\kappa=0.455$
& \underline{91.57}
& \textbf{92.83}
& \textcolor{deepgreen}{\underline{54.05}}
& 74.50
& 75.60
& \textcolor{deepgreen}{56.86}
& 75.41
& \underline{74.24}
& \textbf{83.06}
& $\mathbf{-1.41}$
& \textcolor{deeppurple}{55.46} \\

$\kappa=0.6$
& \textbf{91.60}
& 92.40
& \textcolor{deepgreen}{53.92}
& 73.80
& 75.23
& \textcolor{deepgreen}{\underline{57.23}}
& 75.23
& 74.03
& 82.86
& $\underline{-1.44}$
& \textcolor{deeppurple}{\underline{55.58}} \\

\bottomrule
\end{tabular*}
\end{table*}

\label{sec:kappa-analysis}
\begin{figure}[!t]
\centering
\includegraphics[width=\columnwidth]{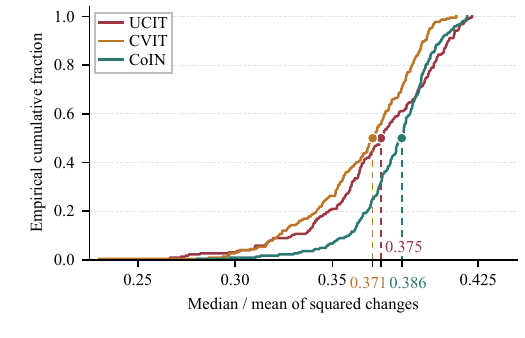}
\caption{Empirical cumulative distributions of module-wise median-to-mean
ratios of squared weight changes on UCIT, CVIT, and CoIN.
Markers and dashed guides indicate the empirical medians.}
\label{fig:median2mean_ratio}
\end{figure}

Fig.~\ref{fig:median2mean_ratio} compares the scale-free change
profiles of the adapted modules. 
Their centers vary across benchmarks, motivating calibration based on the observed changes.
We use the median of the valid module-wise ratios as the empirical
center for each synthesis step. This estimator minimizes the sum of absolute deviations of the
module-wise ratios from the center.
It therefore provides a representative module center for the reference scale in Eq.~(\ref{eq:scale}). Table~\ref{tab:kappa-sensitivity} examines confidence calibration on UCIT.  Here, $\kappa=0$ denotes the confidence-free control; $\kappa=0.375$ matches the observed empirical center; $\kappa=0.455$ is the median-to-mean ratio of squared zero-mean Gaussian variables, whose normalized squares follow a chi-squared distribution with one degree of freedom; and $\kappa=0.6$ tests a larger calibration value. Disabling confidence favors the later Flickr30k task but weakens retention on VizWiz. The Gaussian reference achieves the strongest MAA and BWT, but not the best final or shared-task performance. Further increasing $\kappa$ favors Flickr30k without improving the final average. The empirical-center setting obtains the highest MFN and Sim-Acc,
supporting the use of the empirical center to balance similar tasks.

\subsubsection{Fixed-Rank Restoration}
\begin{table*}[!t]
\caption{Comparison of rank-16 restoration and dense synthesis on
UCIT and CVIT in continual-learning performance and merged LoRA storage.
The best results are in bold.}
\label{tab:svd}
\centering
\scriptsize
\setlength{\tabcolsep}{12pt}
\renewcommand{\arraystretch}{1.08}
\small
\begin{lrbox}{\additionaltablebox}
\begin{tabular}{c|lccccc}
\toprule
Benchmark & Representation & MFT & MFN/AP & MAA/MAP & BWT 
& Merged LoRA Root (M / GiB) \\
\midrule

\multirow{2}{*}{UCIT}
& Rank-16 SVD 
& \textbf{75.54} 
& 74.32
& {83.01} 
& -1.47
& \textbf{39.98 / 0.074} \\

& Dense $\Delta\mathbf{W}$ 
& 75.27 
& \textbf{74.44} 
& \textbf{83.02}
& \textbf{-1.00}
& 6476.01 / 12.062 \\

\hline

\multirow{2}{*}{CVIT}
& Rank-16 SVD 
& \textbf{92.32} 
& \textbf{91.71} 
& \textbf{95.08} 
& -0.74
& \textbf{39.98 / 0.074} \\

& Dense $\Delta\mathbf{W}$ 
& 91.76 
& 91.33 
& 94.66 
& \textbf{-0.51}
& 6476.01 / 12.062 \\

\bottomrule
\end{tabular}
\end{lrbox}
\ifdim\wd\additionaltablebox>\textwidth
\usebox{\additionaltablebox}
\else
\usebox{\additionaltablebox}
\fi
\end{table*}

Table~\ref{tab:svd} compares fixed-rank restoration with retaining the
dense synthesis target. Their final and stage-averaged task performance
remain close, while the low-rank representation requires substantially
less storage. Dense execution has better backward transfer in this
comparison, but does not consistently improve the other task metrics.
The comparison supports fixed-rank restoration as a practical balance
between retained performance and the size of the shared expert.

\begin{figure*}[!t]
\centering
\includegraphics[width=\textwidth]{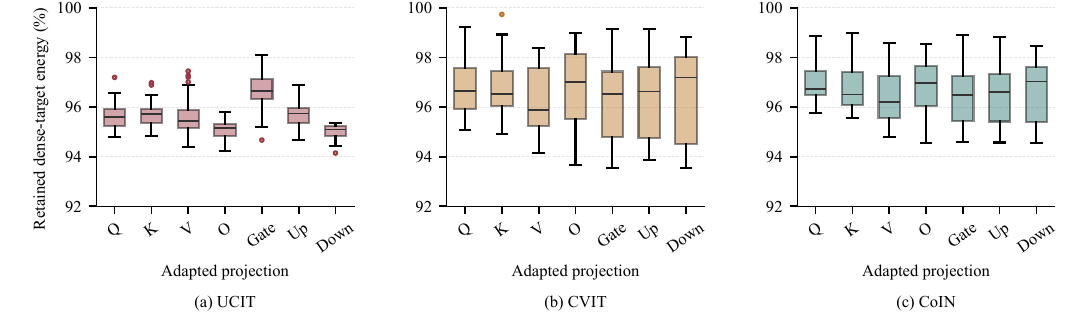}
\caption{Retained synthesis-target energy after fixed-rank restoration
on UCIT, CVIT, and CoIN. Box plots summarize layer-wise results for
each adapted projection.}
\label{fig:rank_restoration}
\end{figure*}

\begin{figure*}[h]
\centering
\includegraphics[width=\textwidth]{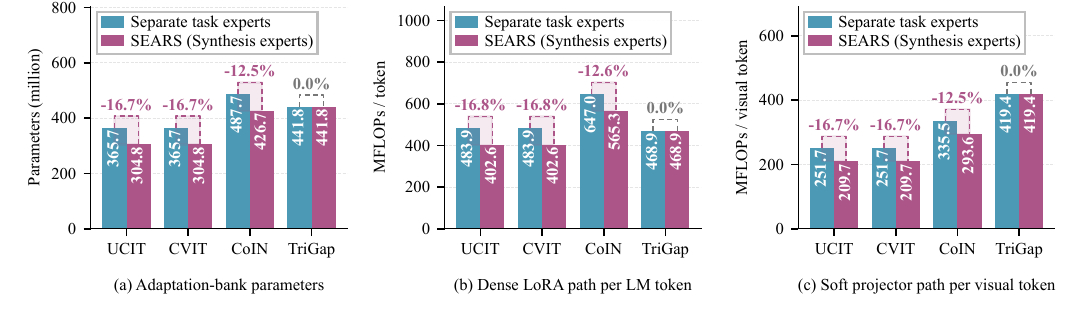}
\caption{
Parameter and computation costs of separate task experts and
\method{} with shared experts. Blue bars retain one expert per task;
purple bars retain and evaluate each logical root once.
(a) Projector and LoRA parameters, (b) dense LoRA computation per
language-model token, and (c) projector computation per visual token.
Dashed brackets indicate relative reductions.
}
\label{fig:export_cost}
\end{figure*}

Fig.~\ref{fig:rank_restoration} shows that fixed-rank restoration preserves most synthesis-target energy across UCIT, CVIT, and CoIN, where a relative Frobenius error $e$ corresponds to retained energy $1-e^2$. Although reconstruction fidelity varies across projections and layers, the dominant components of the synthesized weights are retained. Together with the comparable task performance and substantially lower storage cost in Table~\ref{tab:svd}, this supports fixed-rank restoration as an effective balance between retained performance and adaptation-state size.

\subsubsection{Continued Projector Adaptation}

\begin{figure}[h]
\centering
\includegraphics[width=\columnwidth]{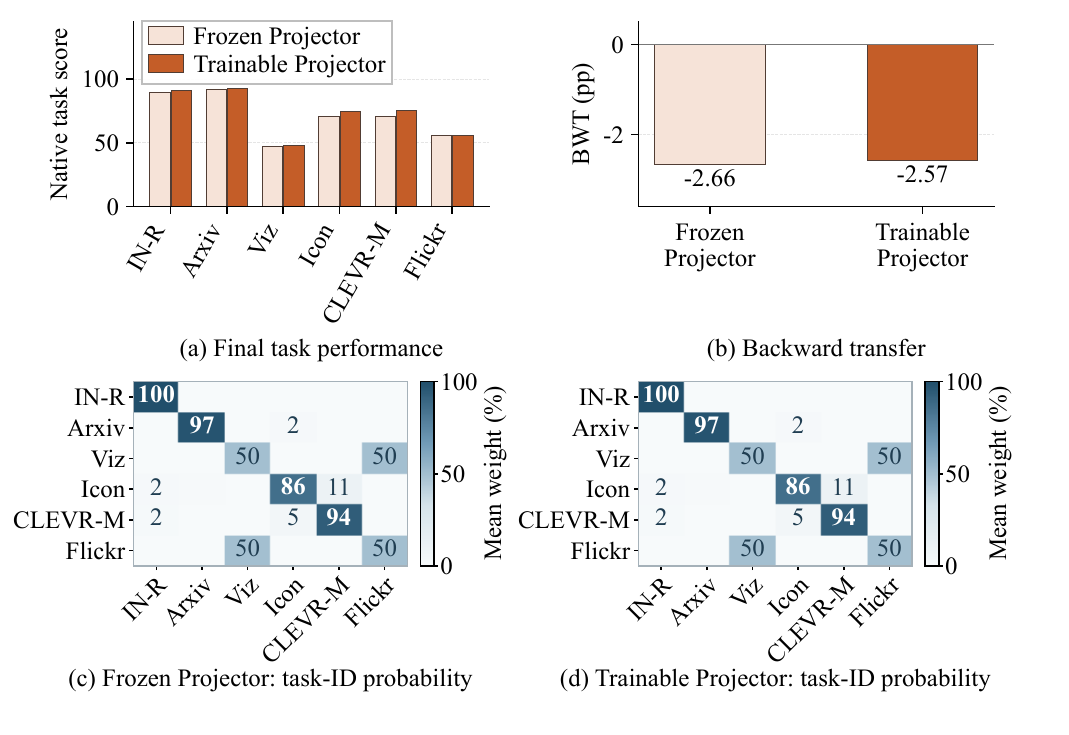}
\caption{Frozen and trainable root-wise projectors on UCIT.
Top: final task performance and backward transfer.
Bottom: task-identity routing distributions.}
\label{fig:mm_adaptation_control}
\end{figure}

Fig.~\ref{fig:mm_adaptation_control} compares frozen and trainable
root-level projectors. For the similar tasks,
continued projector training leaves the task-selection distributions
nearly unchanged and slightly improves final performance, without
worsening aggregate backward transfer. The stable selection patterns
are consistent with instruction scoring being computed independently
by the frozen instruction encoder. These results support continued
projector adaptation for visual--language alignment and retaining the
newly trained projector within a reused root, while applying ARS only
to the effective LoRA weights.

\begin{table*}[h]
\centering
\caption{Comparison with existing methods on UCIT using InternVL.
The best results are in bold, and the second-best results are underlined.}
\label{tab:ucit-internvl}
\setlength{\tabcolsep}{3.5pt}
\renewcommand{\arraystretch}{1.08}
\small
\begin{tabular}{lccccccc|c>{\columncolor{gray!10}}c>{\columncolor{gray!10}}cc}
\toprule
Method & Venue
&  ImageNet-R &  ArXivQA & VizWiz & IconQA & CLEVR-Math & Flickr30k
& MFT$\uparrow$ & MFN$\uparrow$ & MAA$\uparrow$ & BWT$\uparrow$ \\
\midrule
Zero-shot & --
& 21.10 & 63.20 & 40.59 & 24.70 & 21.20 & 44.67
& 35.91 & -- & -- & -- \\

LoRA-FT~\cite{lora2022} & ICLR 2022
& 76.50 & 78.60 & 45.57 & 71.13 & 61.63 & \textbf{58.82}
& 79.20 & 65.38 & 80.66 & -16.59 \\

O-LoRA~\cite{olora2023} & EMNLP 2023
& 90.30 & 93.53 & 46.26 & 68.40 & 65.17 & 56.06
& 70.97 & 69.95 & 81.23 & \underline{-1.22} \\

MoELoRA~\cite{coin2024} & NeurIPS 2024
& 68.57 & 80.47 & 45.43 & 69.23 & 65.93 & 58.57
& \underline{79.25} & 64.70 & 80.07 & -17.46 \\

ModalPrompt~\cite{modalprompt2025} & EMNLP 2025
& 57.10 & 63.90 & 42.11 & 37.40 & 47.50 & 45.76
& 48.98 & 48.96 & 53.52 & \textbf{-0.02} \\

CL-MoE~\cite{clmoe2025} & CVPR 2025
& 74.23 & 78.77 & 44.77 & 53.90 & {72.73} & 58.23
& 71.39 & 63.77 & 77.44 & -9.15 \\

HiDe~\cite{hide2025} & ACL 2025
& 90.00 & {93.77} & {50.49}
& 71.03 & 61.10 & 55.08
& 76.29 & 70.25 & 82.97 & -7.25 \\

SEFE~\cite{sefe2025} & ICML 2025
& 85.93 & 76.90 & 47.42 & 69.77 & 63.37 & 58.19
& 73.55 & 66.93 & 79.44 & -7.94 \\

DISCO~\cite{federated2025} & ICCV 2025
& {92.13} & {94.13} & 48.11
& {73.90} & 67.53 & 58.06
& 78.92 & {72.31} & {84.24} & -7.93 \\

HiDE$^{*}$+PMA~\cite{zhang2026progressive} & ACM MM 2026
& 91.70 & \underline{94.87} & 51.66
& 71.00 & 69.20 & 56.88
& 76.91 & 72.55 & 84.28 & -4.36 \\

DISCO$^{*}$+PMA~\cite{zhang2026progressive} & ACM MM 2026
& \underline{93.37} & \textbf{95.07} & \underline{59.90}
& \underline{74.97} & \underline{82.83} & 58.40
& 79.04 & \underline{77.42} & \underline{86.59} & -1.61 \\

\midrule
\textbf{\method{}} & --
& \textbf{95.23} & 93.67 & \textbf{60.14}
& \textbf{83.57} & \textbf{85.30} &  \underline{58.72}
& \textbf{80.60} & \textbf{79.44} & \textbf{86.94} & -1.39 \\

\bottomrule
\end{tabular}
\end{table*}

\begin{figure}[h]
\centering
\includegraphics[width=0.9\columnwidth]{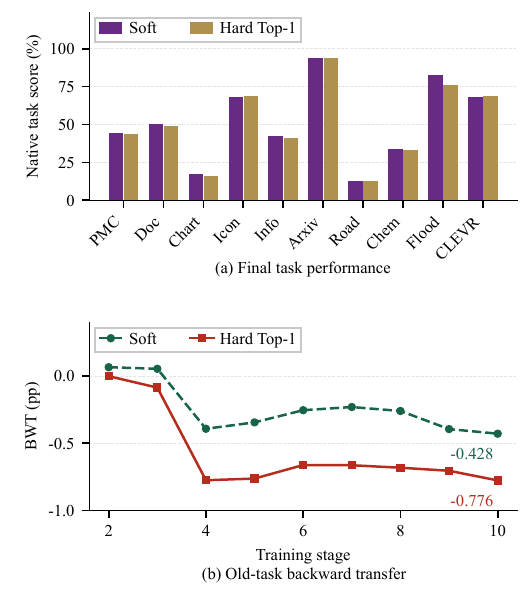}
\vspace{-5mm}
\caption{Soft and hard Top-1 routing on TriGap.
(a) Final task performance and (b) backward transfer over the task stream.}
\vspace{-4mm}
\label{fig:routing_top1}
\end{figure}

\subsection{Computation and Parameter Cost}
\label{sec:cost}
Fig.~\ref{fig:export_cost} examines the storage and computation savings
enabled by expert sharing, using the same backbone dimensions and
per-expert LoRA rank. 
The reference retains one adaptation state per task, whereas
\method{} retains one projector--LoRA state per logical root.
Fig.~\ref{fig:export_cost}(a)
counts projector and LoRA parameters, showing the storage saved by
removing redundant expert copies. Fig.~\ref{fig:export_cost}(b) reports
theoretical FLOPs per language-model token for the dense LoRA block
implementation: fewer expert blocks reduce both factor multiplication
and dense-gate computation. Fig.~\ref{fig:export_cost}(c) reports
projector FLOPs per visual token. Aggregating task probabilities before
projector evaluation allows each shared projector to be computed once.

UCIT, CVIT, and CoIN benefit from shared experts, reducing both
adaptation-parameter storage and inference computation without lowering
the prescribed rank. TriGap instead exhibits more distinct
instruction-conditioned profiles (Fig.~\ref{fig:motivation}), with
observed prototype similarities below the semantic-reuse threshold.
Separate experts are retained in this setting rather than forcing
synthesis across weakly related tasks.

\subsection{Further Analysis}
\label{sec:further-analysis}

\subsubsection{Different Multimodal Backbones}

Table~\ref{tab:ucit-internvl} further evaluates \method{} with
InternVL~\cite{internvl2024}. After replacing LLaVA with InternVL,
\method{} maintains a substantial overall advantage over existing
methods, achieving the strongest MFT, MFN, and MAA. The higher MFT
reflects effective acquisition of new task capabilities, while the
improvements in MFN and MAA show that this advantage persists after
subsequent learning and throughout the task stream. 
These results support the applicability of SEARS across different multimodal backbones.

\subsubsection{Soft and Hard Routing}

Fig.~\ref{fig:routing_top1} compares soft routing with hard Top-1
selection using the same expert bank. Soft routing achieves less
negative backward transfer, indicating better retention of previously
learned capabilities. Unlike hard selection, which relies entirely on
one expert, soft routing preserves probability-weighted contributions
from multiple experts, reducing sensitivity to uncertain expert
selection.

\begin{table*}[h]
\caption{
Sensitivity to LoRA rank on UCIT.
State (M) reports the total projector and LoRA parameters stored
in the five logical roots, in millions.
The best results are boldfaced, and the second-best results are
underlined.
}
\label{tab:rank}
\centering
\scriptsize
\setlength{\tabcolsep}{3pt}
\renewcommand{\arraystretch}{1.12}
\small
\begin{tabular*}{\textwidth}{@{\extracolsep{\fill}}c|c*{11}{c}@{}}
\toprule
Rank
& ImageNet-R & ArXivQA & VizWiz & IconQA & CLEVR-Math & Flickr30k
& MFT$\uparrow$ & MFN$\uparrow$ & MAA$\uparrow$ & BWT$\uparrow$
& State (M) \\
\midrule

 2
& 90.63
& 91.77
& 47.11
& 72.50
& 72.80
& \underline{57.07}
& 74.06
& 71.98
& 81.58
& $-2.49$
& 129.88 \\

 4
& 90.70
& \textbf{93.33}
& 49.95
& 74.40
& 75.13
& 56.72
& 75.12
& 73.37
& 82.50
& $-2.10$
& 154.87 \\

 8
& 91.17
& \underline{93.27}
& 53.28
& 74.30
& 74.77
& 56.69
& 75.04
& 73.91
& \underline{82.81}
& $\mathbf{-1.35}$
& 204.84 \\

 16
& \underline{91.33}
& 92.60
& \underline{54.40}
& \underline{74.63}
& \textbf{76.10}
& 56.84
& \textbf{75.54}
& \textbf{74.32}
& \textbf{83.01}
& $-1.47$
& 304.78 \\

 32
& \textbf{91.37}
& 90.67
& \textbf{54.84}
& \textbf{74.67}
& \underline{75.30}
& \textbf{57.21}
& \underline{75.16}
& \underline{74.01}
& 82.48
& $\underline{-1.38}$
& 504.67 \\

\bottomrule
\end{tabular*}
\end{table*}
\begin{table*}[h]
\centering
\caption{Ten-task class-incremental comparison with and without Gaussian
feature sampling using ViT-B/16-IN21K.
Results for ARS are reported as the mean $\pm$ sample standard
deviation over three random seeds.}
\label{tab:cil-main}
\setlength{\tabcolsep}{4pt}
\renewcommand{\arraystretch}{1.12}
\small
\begin{tabular}{lccccccccccccc}
\toprule
& \multicolumn{2}{c}{10S-ImageNet-R} 
& \multicolumn{2}{c}{10S-ImageNet-A} 
& \multicolumn{2}{c}{10S-CIFAR100} 
& \multicolumn{2}{c}{10S-CUB} \\
Method 
& $A_{\mathrm{Last}} \uparrow $ 
& $A_{\mathrm{Avg}} \uparrow$ 
& $A_{\mathrm{Last}} \uparrow$ 
& $A_{\mathrm{Avg}} \uparrow$ 
& $A_{\mathrm{Last}} \uparrow$ 
& $A_{\mathrm{Avg}} \uparrow$ 
& $A_{\mathrm{Last}} \uparrow$ 
& $A_{\mathrm{Avg}} \uparrow$ \\
\midrule

\multicolumn{9}{c}{\textit{Without Gaussian feature sampling}} \\
\midrule

L2P~\cite{l2p2022} 
& $72.34\pm0.17$ 
& $77.36\pm0.64$ 
& $44.04\pm0.93$ 
& $51.24\pm2.26$ 
& $84.06\pm0.88$ 
& $88.26\pm1.34$ 
& $67.02\pm1.90$ 
& $79.62\pm1.60$ \\

CODA-Prompt~\cite{codaprompt2023} 
& $73.31\pm0.50$ 
& $78.47\pm0.53$ 
& $52.08\pm0.12$ 
& $63.92\pm0.12$ 
& $83.21\pm3.39$ 
& $87.71\pm3.17$ 
& $77.23\pm1.12$ 
& $81.90\pm0.85$ \\

InfLoRA~\cite{inflora2024} 
& $74.75\pm0.64$ 
& $80.67\pm0.55$ 
& $49.20\pm1.12$ 
& $60.92\pm0.61$ 
& $86.75\pm0.35$ 
& $91.72\pm0.15$ 
& $70.82\pm0.23$ 
& $81.39\pm0.14$ \\

SD-LoRA~\cite{sdlora2025} 
& $77.34\pm0.35$ 
& $82.04\pm0.24$ 
& $55.96\pm0.73$ 
& $64.95\pm1.63$ 
& $88.01\pm0.31$ 
& $\mathbf{92.54\pm0.18}$ 
& $77.48\pm0.20$ 
& $85.59\pm0.44$ \\

\textbf{ARS (ours)}
& $\mathbf{79.58\pm0.23}$
& $\mathbf{84.81\pm0.45}$
& $\mathbf{58.22\pm1.13}$
& $\mathbf{66.51\pm1.83}$
& $\mathbf{88.84\pm0.35}$
& ${92.48\pm1.13}$
& $\mathbf{80.52\pm1.24}$
& $\mathbf{88.49\pm0.73}$ \\

\midrule
\midrule

\multicolumn{9}{c}{\textit{With Gaussian feature sampling}} \\
\midrule

SLCA~\cite{slca2023} 
& $79.35\pm0.28$ 
& $83.29\pm0.46$ 
& $61.05\pm0.63$ 
& $68.88\pm2.31$ 
& $91.26\pm0.37$ 
& $94.29\pm0.92$ 
& $84.68\pm0.09$ 
& $90.77\pm0.79$ \\

SSIAT~\cite{ssiat2024} 
& $79.38\pm0.59$ 
& $83.63\pm0.43$ 
& $62.43\pm1.63$ 
& $70.83\pm1.63$ 
& $91.35\pm0.26$ 
& $94.35\pm0.60$ 
& $88.75\pm0.38$ 
& $93.00\pm0.90$ \\

MOS~\cite{mos2025}  
& $77.68\pm0.41$ 
& $82.06\pm0.53$ 
& $54.75\pm1.09$ 
& $63.32\pm2.38$ 
& $\mathbf{91.53\pm0.35}$ 
& $94.21\pm0.91$ 
& $89.97\pm0.32$ 
& $93.43\pm0.60$ \\

\textbf{ARS (ours)}
& $\mathbf{81.65\pm0.45}$
& $\mathbf{86.06\pm0.25}$
& $\mathbf{63.90\pm0.72}$
& $\mathbf{71.07\pm1.75}$
& $\mathbf{91.53\pm0.11}$
& $\mathbf{94.42\pm1.02}$
& $\mathbf{90.35\pm0.11}$
& $\mathbf{93.90\pm0.56}$ \\

\bottomrule
\end{tabular}
\end{table*}
\subsubsection{Sensitivity to LoRA Rank}

Table~\ref{tab:rank} examines the influence of LoRA rank on UCIT.
Increasing the rank initially improves final and stage-averaged
performance, suggesting that insufficient adaptation capacity can
limit learned capabilities. However, the gains diminish as the rank
increases and are not proportional to the added parameters.
Rank $16$ achieves the strongest MFT, MFN, and MAA, whereas further
expansion to rank $32$ increases storage without improving these
metrics. Moreover, a larger rank does not consistently reduce
forgetting. We therefore adopt rank $16$ to balance task performance,
knowledge retention, and parameter cost.

\subsubsection{Class-Incremental Application}
\label{sec:cil-transfer}

Table~\ref{tab:cil-main} evaluates the applicability of ARS to
class-incremental learning on four ten-task benchmarks
using ViT-B/16 pretrained on ImageNet-21K.
The comparisons cover settings with and without Gaussian feature
sampling for classifier adaptation.
Here, ARS consolidates successive LoRA updates without instruction-based expert allocation, and prediction covers all classes learned up to each stage.
L2P~\cite{l2p2022} retrieves input-dependent prompts from a learned
pool, while CODA-Prompt~\cite{codaprompt2023} composes prompt
components through attention.
InfLoRA~\cite{inflora2024} constrains low-rank updates to reduce
inter-task interference, whereas SD-LoRA~\cite{sdlora2025}
decouples the learning of update magnitudes and directions.
SLCA~\cite{slca2023} combines slow representation learning with
classifier alignment.
SSIAT~\cite{ssiat2024} incrementally trains a shared adapter and
compensates for prototype semantic shifts, while
MOS~\cite{mos2025} combines adapter merging with self-refined
adapter retrieval. Without Gaussian feature sampling, ARS achieves the strongest
final accuracy on all four datasets and the strongest stage-averaged
accuracy on three, indicating that ARS remains
effective when consolidating updates across successive classification
tasks. With feature sampling, it maintains the strongest final and stage-averaged accuracy  across all four datasets.
These results suggest that balancing historical and newly acquired
weights helps retain learned capabilities while accommodating new
classes. The consistent advantages across both settings further
support the applicability of ARS beyond multimodal instruction tuning.
\section{Conclusion}
\label{sec:conclusion}

In this work, we presented \method{} for multimodal continual instruction
tuning. It retains task-specific instruction prototypes while selectively
sharing projector--LoRA experts. Specifically, Semantic Expert Allocation
uses instruction similarity to determine whether an arriving task should
reuse an existing logical root or create a new one. For a reused root,
post-training Adaptive Retention Synthesis (ARS) combines its stored and
newly acquired effective LoRA weights using relative weight amplitudes
and empirically calibrated change confidence. Truncated SVD restores
the prescribed LoRA rank, and the synthesis requires neither additional
learnable parameters nor stored importance statistics. Extensive
experiments on four benchmarks demonstrate leading overall performance
and improved retention on previously learned tasks. Further evaluations
support the applicability of \method{} across multimodal backbones and
the effectiveness of ARS in class-incremental learning.

\noindent\textit{Limitations \& Future Directions:}
Our current framework adopts a fixed semantic threshold and
prescribed LoRA ranks. Future work will explore more flexible expert
allocation and rank selection, and extend the framework to longer
task streams and more diverse application scenarios.

\bibliographystyle{IEEEtran}
\bibliography{references}
\clearpage
\appendices

This supplement provides the complete algorithm and
further analysis for the main paper, with references
in \textcolor{red}{red} following the main paper's numbering.

\section{Notation and Complete Procedure}
\label{app:notation-summary}

Table~\ref{tab:principal-notation} summarizes the principal quantities in
\method{}. Algorithm~\ref{alg:complete} follows the task stream from
semantic allocation and current-task training to ARS and root-level
inference. A separate prototype is stored for every arriving task,
whereas synthesis is applied when the selected expert is reused.
\begin{table*}[!t]
\centering
\caption{Principal notation and parameter roles in \method{}.}
\label{tab:principal-notation}
\footnotesize
\setlength{\tabcolsep}{5pt}
\renewcommand{\arraystretch}{1.04}
\begin{tabular}{@{}p{0.30\textwidth}p{0.67\textwidth}@{}}
\toprule
Symbol & Description \\
\midrule
\multicolumn{2}{@{}l}{\textit{Task stream and adaptation state}}\\
$T,t;\mathcal{D}_t,N_t$
& Number of tasks, current stage, current-task training set, and its sample count.\\
$\mathbf{X}_{t,n},\mathbf{u}_{t,n},\mathbf{y}_{t,n}$
& Image, instruction, and target response of example $n$ in task $t$.\\
$N_{t,n}^{\rm ins},N_{t,n}^{\rm ans}$
& Instruction and supervised response lengths (in tokens) of that example.\\
$\mathcal{G}_t,K_t;\mathcal{H}_t$
& Reachable roots, their count, and the task-to-root assignment after stage $t$.\\
$j_t^\star,g_t,\iota_t$
& Most similar historical identity, selected root, and indicator of accepted reuse.\\
$L;D_{{\rm in},\ell},D_{{\rm out},\ell}$
& Number of adapted mappings and the input/output widths of mapping $\ell$.\\
$\mathbf{W}_{0,\ell},\Delta\mathbf{W}_{g,\ell}$
& Frozen backbone weight and root-$g$ effective LoRA update in mapping $\ell$.\\
$\mathbf{A}_{g,\ell},\mathbf{B}_{g,\ell};R_\ell,\varsigma_\ell$
& LoRA factors, prescribed rank, and fixed scale, with
$\Delta\mathbf{W}_{g,\ell}=\varsigma_\ell\mathbf{B}_{g,\ell}\mathbf{A}_{g,\ell}$.\\
$\mathcal{P}_g,\boldsymbol{\Phi}_g,\boldsymbol{\theta}_g$
& Root-specific projector, its parameter vector, and the joint trainable vector of projector parameters and LoRA factors.\\
$\mathbf{X}^{\rm vis};N_v,D_v,D_m$
& Frozen visual-token features, token count, visual-feature width, and language-model width.\\
\midrule
\multicolumn{2}{@{}l}{\textit{Semantic allocation, soft inference, and performance analysis}}\\
$\mathcal{F}_{\rm txt},D_c;\operatorname{Norm}$
& Frozen instruction encoder, feature dimension, and unit-$\ell_2$ normalization.\\
$N_t^{\rm scan},\mathbf{c}^{\rm scan}_t,\mathbf{c}^{\rm store}_t$
& Pre-training scan size, normalized scan prototype, and stored task prototype.\\
$s_{t,j},\Theta$
& Prototype similarity and reuse threshold; reuse is accepted when $s_{t,j_t^\star}\geq\Theta$.\\
$\mathcal{L}_t,\mathcal L_a^{\rm eval};p_{\mathcal M}$
& Token-averaged training and evaluation NLLs and the model's conditional next-token distribution.\\
$E,K$
& Identity and root counts at evaluation: $E=t$, $K=K_t$; at the final stage, $E=T$.\\
$\mathbf{f},\mathbf{s},s_j,T_r$
& Test-instruction feature, identity-score vector, its scalar entries, and positive softmax temperature.\\
$\mathbf{G}_t$
& Binary root-membership matrix in $\{0,1\}^{E\times K}$, with $G_{t,jk}=\mathbb{I}[\mathcal{H}_t(j)=r_k]$.\\
$\boldsymbol\pi^{\rm task},\boldsymbol\pi^{\rm root}$
& Instruction-dependent routing weights over task identities and roots, with
$\boldsymbol\pi^{\rm root}=\mathbf G_t^\top\boldsymbol\pi^{\rm task}$ and scalar entries $\pi_j^{\rm task}$ and $\pi_k^{\rm root}$.\\
$\Delta\mathbf{W}^{\rm route}_\ell,\mathbf{O}^{\rm soft}$
& Root-weighted effective LoRA update and mixture of projector outputs at inference.\\
$\xi,J,\widehat J;\mathcal{S},\mathcal{C}$
& Evaluation example, hidden and dominant predicted identities; response-success and correct-root-association events.\\
$\Pr(\cdot)$
& Probability of an event over the evaluation distribution; distinct from per-instruction routing weights.\\
\midrule
\multicolumn{2}{@{}l}{\textit{Adaptive Retention Synthesis and empirical calibration}}\\
$\Delta\mathbf{W}^{\rm old},\Delta\mathbf{W}^{\rm new}$
& Stored and current-task-trained effective LoRA weights of the reused root.\\
$\mathbf{D},\mathbf{Z}$
& Current-task displacement $\Delta\mathbf{W}^{\rm new}-\Delta\mathbf{W}^{\rm old}$ and its entry-wise square $\mathbf{D}^{\odot2}$.\\
$\alpha,\varepsilon$
& Amplitude exponent ($\alpha=2$) and positive numerical-stability constant.\\
$\mathbf{I}^{\rm old},\mathbf{I}^{\rm new},\boldsymbol{\Lambda}^{\rm amp}$
& Weight-amplitude importance of the two states and the proposed new-state share.\\
$\mathcal{U}_t,\mathcal{U}_t^{+}$
& Participating mappings and the subset with positive squared-change median and mean.\\
$n_\ell,m_{t,\ell},\bar z_{t,\ell}$
& Module entry count, median squared change, and mean squared change.\\
$\operatorname{Med}_{\rm lo},\operatorname{Med}_{\rm mid}$
& Median operators using, for an even count, the lower central entry and the average of the two central values, respectively.\\
$\eta_{t,\ell},\widehat F_t$
& Ratio $m_{t,\ell}/\bar z_{t,\ell}$ and its equally weighted empirical CDF over $\mathcal{U}_t^{+}$.\\
$\kappa_t$
& Current-synthesis calibration: the midpoint median of $\{\eta_{t,\ell}:\ell\in\mathcal{U}_t^{+}\}$, computed before write-back.\\
$\tau_{t,\ell}^{2},\chi_{t,\ell}$
& Raw reference $m_{t,\ell}/\kappa_t$ and stabilized denominator scale $\max\{\tau_{t,\ell}^{2},\varepsilon\}+\varepsilon$.\\
$\rho_{t,\ell,ab},\boldsymbol{\rho}_{t,\ell}$
& Entry-wise update-retention coefficient and its matrix form.\\
$\mathbf D_\ell^{\rm amp},\mathbf D_\ell^{\rm ret}$
& Amplitude-proposed and retained displacements, with
$\mathbf D_\ell^{\rm ret}=\boldsymbol\rho_\ell\odot\mathbf D_\ell^{\rm amp}$.\\
$\boldsymbol{\Lambda},\Delta\mathbf{W}^{\star}$
& Final mixing matrix $\boldsymbol{\rho}\odot\boldsymbol{\Lambda}^{\rm amp}$ and dense synthesis target $\Delta\mathbf{W}^{\rm old}+\boldsymbol{\Lambda}\odot\mathbf{D}$.\\
$\kappa_{\rm G}$
& Gaussian-reference ratio used only for theoretical and sensitivity comparisons, not as the adaptive calibration value.\\
\midrule
\multicolumn{2}{@{}l}{\textit{Fixed-rank restoration, stability, and storage}}\\
$\mathcal{J}(\mathbf{M}),C_\Lambda$
& Weighted reconstruction objective for fixed $\boldsymbol{\Lambda}$ and its additive constant after completing the square.\\
$\mathbf{U},\boldsymbol{\Sigma},\mathbf{V};\sigma_i$
& Compact-SVD factors of the dense target and its singular values.\\
$d_{\min},\bar R$
& Compact-SVD width $\min(D_{\rm out},D_{\rm in})$ and retained rank $\min(R,d_{\min})$.\\
$\widehat{\Delta\mathbf{W}},e_R$
& Rank-restored update and residual norm, with $e_R^2=\sum_{i>\bar R}\sigma_i^2$.\\
$\mathbf{A}^{\star},\mathbf{B}^{\star}$
& Restored factors satisfying $\varsigma\mathbf{B}^{\star}\mathbf{A}^{\star}=\widehat{\Delta\mathbf{W}}$.\\
$D_{\rm LoRA},D_{\rm proj},\Delta D_{\rm bank}$
& LoRA-factor and projector parameter counts per root; storage saving $(E-K)(D_{\rm LoRA}+D_{\rm proj})$.\\
$\boldsymbol{\vartheta},D_{\rm eff}$
& Effective-weight/projector coordinates used for stability analysis, with $D_{\rm eff}=\sum_\ell D_{{\rm out},\ell}D_{{\rm in},\ell}$; distinct from trainable factor coordinates $\boldsymbol{\theta}_g$.\\
$\ell_{\rm resp},\Gamma_\ell(\xi),\Gamma_\Phi(\xi),\Gamma_\pi(\xi)$
& Per-example response loss and local gradient bounds with respect to the
effective LoRA weights, projector parameters, and root routing weights.\\
$\odot,\oslash,\|\cdot\|_F$
& Entry-wise product, entry-wise division, and Frobenius norm.\\
\bottomrule
\end{tabular}
\end{table*}

\begin{algorithm}[!t]
\caption{Complete Procedure of \method{}}
\label{alg:complete}
\footnotesize
\begin{algorithmic}[1]
\REQUIRE Task stream $\{\mathcal D_t\}_{t=1}^{T}$; frozen encoders and
backbone; threshold $\Theta$; ranks $R_\ell$ and scales $\varsigma_\ell$;
exponent $\alpha$; positive floor $\varepsilon$
\ENSURE Task prototypes, assignment $\mathcal H_T$, and root adaptation states
\STATE Initialize $\mathcal G_0=\varnothing$ and empty $\mathcal H_0$
\FOR{$t=1,\ldots,T$}
  \STATE Compute $\mathbf c_t^{\rm scan}$ by Eq. \textcolor{red}{(3)}
  \STATE Select $g_t$ by Eq. \textcolor{red}{(5)}; extend $\mathcal H_{t-1}$
  with $\mathcal H_t(t)=g_t$
  \IF{$g_t=t$}
    \STATE Initialize the projector from the pretrained multimodal
    projector, $\mathbf A$ with Gaussian weights, and $\mathbf B$ with zeros
  \ELSE
    \STATE Save the selected root's stored LoRA factors and effective weights
    \STATE Initialize its projector and LoRA factors from the latest stored state
  \ENDIF
  \STATE Train only root $g_t$ on $\mathcal D_t$ by Eq. \textcolor{red}{(6)}
  \STATE Store the normalized mean of instruction features accumulated
  during training as $\mathbf c_t^{\rm store}$
  \IF{$g_t\ne t$}
    \STATE Collect all module change statistics and form $\mathcal U_t^+$
    by Eqs. \textcolor{red}{(11)}--\textcolor{red}{(12)}
    \IF{$\mathcal U_t^+\ne\varnothing$}
      \STATE Compute $\kappa_t$ by Eq. \textcolor{red}{(12)}
      \FOR{each mapping $\ell\in\mathcal U_t$}
        \STATE Compute the amplitude share, reference scale, and
        retention coefficient by Eqs. \textcolor{red}{(10)},
        \textcolor{red}{(13)}, and \textcolor{red}{(15)}
        \STATE Form $\Delta\mathbf W_{g_t,\ell}^{\star}$ by Eq. \textcolor{red}{(16)}
        \STATE Truncate the SVD to $\bar R_\ell$ components and recover the
        factors by Eq.~(\ref{eq:lora-factor-recovery})
      \ENDFOR
    \ELSE
      \STATE Restore the stored LoRA factors
    \ENDIF
  \ENDIF
  \STATE Retain the acquired projector and update reachable roots $\mathcal G_t$
\ENDFOR
\STATE Compute task routing weights by Eq. \textcolor{red}{(21)}
\STATE Aggregate routing weights by Eq. \textcolor{red}{(22)}
\STATE Execute Eqs. \textcolor{red}{(23)}--\textcolor{red}{(24)}
\end{algorithmic}
\end{algorithm}

\section{Supplementary Analysis of \method{}}
\label{app:supplementary-derivations}

We first analyze semantic allocation and root-level execution, followed
by the weight-magnitude proxy and fixed-rank reconstruction. We then
develop a unified analysis of empirical calibration, update retention,
and response variation. The final subsection examines shared-expert
adaptation through paired LoRA and projector comparisons.

\subsection{Semantic Allocation and Root-Level Execution}
\label{app:probability-proof}

\paragraph{Task association and response success}
At stage $t$, let $\mathcal S$ and $\mathcal C$ be the response-success
and correct-root-association events defined in
Section.\textcolor{red}{III-C (Inter- and Intra-Task Performance)}. In particular,
\begin{equation}
\mathcal C=\{\mathcal H_t(\widehat J)=\mathcal H_t(J)\}.
\label{eq:app-root-association}
\end{equation}
The disjoint events $\mathcal S\cap\mathcal C$ and
$\mathcal S\cap\mathcal C^c$ give
\begin{equation}
\begin{aligned}
&\Pr(\mathcal S)=\Pr(\mathcal S\cap\mathcal C)
+\Pr(\mathcal S\cap\mathcal C^c),\\
&\Pr(\mathcal S\cap\mathcal C)
=\Pr(\mathcal C)\Pr(\mathcal S\mid\mathcal C).
\end{aligned}
\label{eq:app-total-probability}
\end{equation}
Applying the same product rule to the second joint term gives
Eq. \textcolor{red}{(2)}. A stratum with zero
probability contributes zero. The contribution of responses outside
the correct association is bounded by
\begin{equation}
0\leq\Pr(\mathcal S)-\Pr(\mathcal S\cap\mathcal C)
=\Pr(\mathcal S\cap\mathcal C^c)\leq\Pr(\mathcal C^c).
\label{eq:app-product-remainder}
\end{equation}
Correct task prediction implies correct root association, so
\begin{equation}
\begin{aligned}
\Pr(\mathcal C)-\Pr(\widehat J=J)
&=\Pr(\{\widehat J\ne J\}\cap\mathcal C)\geq0.
\end{aligned}
\label{eq:quotient-correctness}
\end{equation}
Thus an identity mismatch within a shared root does not change which
expert the two identities refer to.

\paragraph{Exact grouped execution}
For an instruction $\mathbf u$, group the expert operators
$\mathcal A_g$ by their root assignment:
\begin{equation}
\begin{aligned}
&\sum_{j=1}^{E}\pi_j^{\rm task}(\mathbf u)\mathcal A_{\mathcal H_t(j)}\\
&\quad=\sum_{g\in\mathcal G_t}
\left(\sum_{j:\mathcal H_t(j)=g}\pi_j^{\rm task}(\mathbf u)\right)\mathcal A_g.
\end{aligned}
\label{eq:app-exact-root-grouping}
\end{equation}
Applying this identity to effective LoRA matrices and to separately
evaluated projector outputs yields
Eqs. \textcolor{red}{(23)}--\textcolor{red}{(24)}. The shared
expert is evaluated once with its aggregated probability. Its contribution
depends on the total mass assigned to that root, while task prototypes
retain the instruction distinctions used to compute the identity scores.

\paragraph{Instruction similarity and selective reuse}
\label{sec:similarity-reuse}
With $T_r=0.05$, the task routing weight in
Eq. \textcolor{red}{(21)} is
\begin{equation}
\pi_j^{\rm task}(\mathbf u)=
\frac{\exp(s_j(\mathbf u)/T_r)}
{\sum_{i=1}^{E}\exp(s_i(\mathbf u)/T_r)}.
\label{eq:routing-softmax-analysis}
\end{equation}
For evaluation task $a$, let $\mathbf{u}_{a,n}^{\rm eval}$ be the
instruction in example $n$. Its mean routing weight on identity $j$ is
$\bar\pi_{a\to j}^{\rm task}
=(N_a^{\rm eval})^{-1}\sum_{n=1}^{N_a^{\rm eval}}
\pi_j^{\rm task}(\mathbf u_{a,n}^{\rm eval})$.
For two candidate identities $a,b$, define the per-instruction share
\begin{equation}
\begin{aligned}
\pi^{\rm pair}_{a\mid a,b}(\mathbf u)
&=\frac{\pi_a^{\rm task}(\mathbf u)}
{\pi_a^{\rm task}(\mathbf u)+\pi_b^{\rm task}(\mathbf u)}\\
&=\frac{1}{1+\exp[-(s_a(\mathbf u)-s_b(\mathbf u))/T_r]}.
\end{aligned}
\label{eq:pair-normalized-routing}
\end{equation}
A score gap no greater than one temperature unit corresponds to
\begin{equation}
\begin{aligned}
|s_a(\mathbf u)-s_b(\mathbf u)|\leq T_r
&\Longleftrightarrow \pi^{\rm pair}_{a\mid a,b}\in\mathcal I_{\rm amb},\\
\mathcal I_{\rm amb}
&=\left[\frac{1}{1+e},\frac{e}{1+e}\right].
\end{aligned}
\label{eq:routing-ambiguity-interval}
\end{equation}
For a query equal to a normalized task prototype, its self-score is
one. In this case, the interval corresponds to a competing score in
$[1-T_r,1]=[0.95,1]$.

\begin{table*}[!t]
\centering
\caption{Instruction similarity and mean task-identity routing
probabilities before root aggregation.
$s_{\rm alloc}$ denotes the allocation similarity of the task pair.
Column $\bar\pi_j$ reports the mean task routing weight
$\bar\pi_{a\to j}^{\rm task}$ (in percent) for evaluation task $a$.
The order of task identities $j$ is given for each benchmark.
Bold entries correspond to the paired tasks.}
\label{tab:similarity-routing}
\begingroup
\small
\setlength{\tabcolsep}{4pt}
\renewcommand{\arraystretch}{1.08}
\begin{tabular}{@{}llc*{6}{r}@{}}
\toprule
Benchmark & Evaluation task & $s_{\rm alloc}$
& $\bar\pi_1$ & $\bar\pi_2$ & $\bar\pi_3$
& $\bar\pi_4$ & $\bar\pi_5$ & $\bar\pi_6$ \\
\midrule
\multicolumn{9}{@{}l}{
\textit{UCIT: ImageNet-R, ArXivQA, VizWiz, IconQA,
CLEVR-Math, Flickr30k.}}\\
UCIT & VizWiz
& 0.980
& 0.029 & 0.000 & \textbf{49.985}
& 0.000 & 0.000 & \textbf{49.985} \\
UCIT & Flickr30k
& 0.980
& 0.029 & 0.000 & \textbf{49.985}
& 0.000 & 0.000 & \textbf{49.985} \\
\midrule
\multicolumn{9}{@{}l}{
\textit{CVIT: ScienceQA, TextVQA, Flickr30k, ImageNet,
GQA, VQAv2.}}\\
CVIT & GQA
& 0.963
& 1.712 & 0.893 & 0.566
& 5.036 & \textbf{49.541} & \textbf{42.252} \\
CVIT & VQAv2
& 0.963
& 0.722 & 0.583 & 1.059
& 5.954 & \textbf{41.889} & \textbf{49.793} \\
\bottomrule
\end{tabular}
\endgroup
\end{table*}

\providecolor{pairedtask}{RGB}{0,100,60}
\providecolor{pairedscore}{RGB}{128,70,170}
\begin{table*}[!t]
\centering
\caption{Closest-pair analysis on TriGap after the fifth task.
(a) Mean routing probabilities over the five learned task experts
(percentages). (b) Performance with separate experts and forced fusion
of DocVQA and InfographicVQA. $K_5$ is the logical-root count;
Pair Avg. averages the two task scores, and Stream Avg. averages
all five learned tasks. Best and second-best performance is
boldfaced and underlined.}
\label{tab:trigap-reuse-control}
\begingroup
\small
\setlength{\tabcolsep}{4pt}
\renewcommand{\arraystretch}{1.08}
\textit{(a) Allocation similarity and normalized routing probabilities}\\[3pt]
\begin{tabular}{@{}lc c>{\color{pairedtask}}c cc>{\color{pairedtask}}c@{}}
\toprule
Evaluation task & $s_{\rm alloc}$
& PMCVQA & \multicolumn{1}{c}{DocVQA}
& ChartQA & IconQA & \multicolumn{1}{c}{InfographicVQA} \\
\midrule
DocVQA & 0.843
& 0.349 & \textbf{71.825} & 1.285 & 2.119 & 24.421 \\
InfographicVQA & 0.843
& 0.618 & 22.321 & 5.500 & 2.204 & \textbf{69.359} \\
\bottomrule
\end{tabular}

\medskip
\textit{(b) Separate experts and forced fusion}\\[3pt]
\begin{tabular}{@{}lcc >{\color{pairedtask}}c
>{\color{pairedtask}}c >{\color{pairedscore}}c
>{\columncolor{gray!10}}c@{}}
\toprule
Setting & $s_{\rm alloc}$ & $K_5$
& \multicolumn{1}{c}{DocVQA$\uparrow$}
& \multicolumn{1}{c}{InfographicVQA$\uparrow$}
& \multicolumn{1}{c}{Pair Avg.$\uparrow$}
& Stream Avg.$\uparrow$ \\
\midrule
Separate experts & 0.843 & 5
& \textbf{49.99} & \textbf{42.27} & \textbf{46.13} & \textbf{44.71} \\
Forced fusion & 0.843 & 4
& \underline{48.13} & \underline{41.85}
& \underline{44.99} & \underline{44.60} \\
\bottomrule
\end{tabular}
\endgroup
\end{table*}

Table~\ref{tab:similarity-routing} illustrates this probability sharing.
VizWiz and Flickr30k have allocation similarity $0.980$ and receive
approximately equal identity probabilities; their shared root receives
$99.970\%$ of the total mass. On CVIT, the GQA--VQAv2 similarity is
$0.963$, and their pair-normalized mean shares lie within
$\mathcal I_{\rm amb}$. The reported pair share is
$\bar\pi_{a\to a}^{\rm task}/
(\bar\pi_{a\to a}^{\rm task}+\bar\pi_{a\to b}^{\rm task})$,
normalizing after averaging each identity's routing weights.
Root aggregation therefore combines competing identity probabilities
when the corresponding tasks share an executable expert.

TriGap presents a different allocation pattern. Its closest pair,
DocVQA and InfographicVQA, has similarity $0.843$ and own-task routing
probabilities $71.825\%$ and $69.359\%$ after the fifth task. Their
pair-normalized own-task shares, $0.746$ and $0.757$, lie outside
$\mathcal I_{\rm amb}$. In Table~\ref{tab:trigap-reuse-control}, forcing
InfographicVQA to initialize from DocVQA and applying synthesis reduces
the pair average from $46.13$ to $44.99$. Keeping separate experts
preserves their distinct task requirements, supporting selective rather
than unconditional expert reuse.

\subsection{Weight-Magnitude Importance Proxy}
\label{app:amplitude-proof}

\paragraph{Effective weights and connection contribution}
For an invertible matrix $\mathbf C$, the two LoRA factorizations obey
\begin{equation}
(\mathbf B\mathbf C)(\mathbf C^{-1}\mathbf A)=\mathbf B\mathbf A.
\label{eq:app-factor-invariance}
\end{equation}
Consequently, computing synthesis statistics from the effective weight
makes them invariant to equivalent choices of LoRA factors. For one
mapping and state $\zeta\in\{\mathrm{old},\mathrm{new}\}$, let
$w_{ij}^{\zeta}=(\Delta\mathbf W^{\zeta})_{ij}$ be the effective weight
connecting input channel $j$ to output channel $i$, and let
$x_j^{\zeta}$ be the activation of input channel $j$. Define
\begin{equation}
\mu_j^{\zeta}=\mathbb E|x_j^{\zeta}|,\qquad
\nu_j^{\zeta}=\mathbb E[(x_j^{\zeta})^2].
\label{eq:app-activation-moments}
\end{equation}
Dohare et al.~\cite{dohare2024plasticity} estimate connection contribution
using $|w_{ij}^{\zeta}|\mu_j^{\zeta}$ and sum over outgoing connections
for unit utility. With the weight fixed, the squared response satisfies
\begin{equation}
\mathbb E[(w_{ij}^{\zeta}x_j^{\zeta})^2]
=(w_{ij}^{\zeta})^2\nu_j^{\zeta}.
\label{eq:app-connection-energy}
\end{equation}
When the input second moments agree, their common factor cancels from
the relative connection energy, motivating the squared-weight proxy.

\paragraph{Activation agreement}
We estimate channel moments using
deterministic preprocessing and valid tokens. For task $u$ and mapping
$\ell$, let $\boldsymbol\mu_\ell^u$ collect the mean absolute channel
activations. The profile similarity is
\begin{equation}
c_\ell(u,v)=\frac{\langle\boldsymbol\mu_\ell^u,
\boldsymbol\mu_\ell^v\rangle}
{\|\boldsymbol\mu_\ell^u\|_2\|\boldsymbol\mu_\ell^v\|_2}.
\label{eq:app-activation-profile-cosine}
\end{equation}
Table~\ref{tab:activation-statistics} summarizes the module-averaged
statistics. UCIT acquisition-state activation magnitudes range from
$0.187464$ to $0.193432$. With the same acquired expert, VizWiz and
Flickr30k have mean absolute activations $0.192823$ and $0.193432$,
profile cosine $0.996136$, and mean channel-magnitude agreement
$1-|\mu_j^{\rm Flickr30k}-\mu_j^{\rm VizWiz}|/
(\mu_j^{\rm Flickr30k}+\mu_j^{\rm VizWiz})$ of $0.967768$.

For ten-task CUB, we evaluate all tasks with a fixed final model for
each seed and average over three seeds. The 45 task pairs have mean
profile cosine $0.999573$, with pair means between $0.999333$ and
$0.999750$; task activation magnitudes range from $0.656017$ to
$0.658026$. The agreement of activation directions and scales supports
estimating relative connection contribution from weight magnitudes
without retaining task-wise activation statistics in the expert.
\begin{table*}[!t]
\centering
\caption{Activation statistics supporting the weight-magnitude proxy.
(a) UCIT acquisition states, averaged over valid tokens and input channels.
(b) CUB task profiles under a fixed final model per seed, without Gaussian
feature sampling; values average three seeds, and SD is the sample
standard deviation. (c) Profile similarity and RMS mismatch.
Channel means weight all input channels equally; module means weight
adapted matrices equally. Each CUB task's pair cosine averages its nine
partners. RMS discrepancies are averaged over channels within each
module and then over modules.}
\label{tab:activation-statistics}
\begingroup
\small
\setlength{\tabcolsep}{4pt}
\renewcommand{\arraystretch}{1.04}
\begin{minipage}[t]{0.47\textwidth}
\centering
\textit{(a) UCIT mean absolute activations}\\[3pt]
\begin{tabular}{@{}lrr@{}}
\toprule
Task & Channel mean & Module mean \\
\midrule
ImageNet-R & 0.161395 & 0.189794 \\
ArXivQA & 0.161331 & 0.189792 \\
VizWiz & 0.164489 & 0.192971 \\
IconQA & 0.163729 & 0.192753 \\
CLEVR-Math & 0.159162 & 0.187464 \\
Flickr30k & 0.164943 & 0.193432 \\
\bottomrule
\end{tabular}
\end{minipage}\hfill
\begin{minipage}[t]{0.51\textwidth}
\centering
\textit{(b) CUB mean absolute activations}\\[3pt]
\begin{tabular}{@{}crr@{}}
\toprule
Task & Module mean $\pm$ SD & Mean pair cosine \\
\midrule
1 & $0.657097\pm0.003144$ & 0.999560 \\
2 & $0.658026\pm0.002041$ & 0.999440 \\
3 & $0.656591\pm0.002515$ & 0.999606 \\
4 & $0.656979\pm0.000986$ & 0.999666 \\
5 & $0.656422\pm0.002608$ & 0.999559 \\
6 & $0.656017\pm0.001820$ & 0.999536 \\
7 & $0.656618\pm0.001533$ & 0.999580 \\
8 & $0.657201\pm0.002450$ & 0.999558 \\
9 & $0.656163\pm0.000682$ & 0.999640 \\
10 & $0.657336\pm0.001247$ & 0.999585 \\
\bottomrule
\end{tabular}
\end{minipage}
\medskip

\textit{(c) Mean-absolute profile cosine and channel-wise RMS mismatch}\\[3pt]
\begin{tabular}{@{}lrr@{}}
\toprule
Comparison & Mean cosine & Mean $\delta^{\rm rms}$ \\
\midrule
UCIT: common-state VizWiz--Flickr30k
& 0.996136 & 0.033585 \\
UCIT: stored--acquired activation states
& 0.985933 & 0.055722 \\
CUB: 45 task pairs
& 0.999573 & 0.010263 \\
\bottomrule
\end{tabular}
\endgroup
\end{table*}

\paragraph{Approximation error}
Set $a=(|w_{ij}^{\rm old}|+\varepsilon)^2>0$,
$b=(|w_{ij}^{\rm new}|+\varepsilon)^2>0$, and $S=a+b$.
The weight-only and activation-aware energy shares are
\begin{equation}
\lambda_{ij}^{0}=\frac{b}{a+b},\qquad
\lambda_{ij}^{\rm act}=
\frac{b\nu_j^{\rm new}}{a\nu_j^{\rm old}+b\nu_j^{\rm new}}.
\label{eq:app-activation-energy-shares}
\end{equation}
Their discrepancy is bounded by the mismatch of input RMS magnitudes.

\begin{proposition}[Weight-magnitude approximation]
\label{prop:activation-proxy}
Suppose $\nu_j^{\rm old},\nu_j^{\rm new}>0$ and define
\begin{equation}
\delta_j^{\rm rms}=
\frac{|\sqrt{\nu_j^{\rm new}}-\sqrt{\nu_j^{\rm old}}|}
{\sqrt{\nu_j^{\rm new}}+\sqrt{\nu_j^{\rm old}}}.
\label{eq:app-rms-discrepancy}
\end{equation}
Then
\begin{equation}
|\lambda_{ij}^{\rm act}-\lambda_{ij}^{0}|\leq\delta_j^{\rm rms}.
\label{eq:app-activation-share-bound}
\end{equation}
For an $m\times n$ mapping, this gives
\begin{equation}
\frac{1}{mn}\sum_{i,j}|\lambda_{ij}^{\rm act}-\lambda_{ij}^{0}|
\leq\frac1n\sum_j\delta_j^{\rm rms}.
\label{eq:app-mean-activation-bound}
\end{equation}
\end{proposition}

\paragraph{Proof}
Write $r_o=\sqrt{\nu_j^{\rm old}}$ and $r_n=\sqrt{\nu_j^{\rm new}}$.
Subtracting the shares yields
\begin{equation}
|\lambda_{ij}^{\rm act}-\lambda_{ij}^{0}|
=\frac{ab|r_n^2-r_o^2|}{(a+b)(ar_o^2+br_n^2)}.
\end{equation}
Since
\begin{equation}
(a+b)(ar_o^2+br_n^2)-ab(r_o+r_n)^2=(ar_o-br_n)^2\geq0,
\end{equation}
the ratio is at most $|r_n-r_o|/(r_n+r_o)$. Averaging over output
and input indices proves the second inequality.\hfill$\square$

The mean RMS discrepancies are $0.033585$ for the common-state UCIT
pair, $0.055722$ for the stored/acquired states in the activation-weighted
control, and $0.010263$ for CUB pairs. For these empirical moments,
Eq.~(\ref{eq:app-mean-activation-bound}) bounds the corresponding share
errors averaged within each mapping and then equally over mappings.

The implemented amplitude share is $\Lambda_{ij}^{\rm amp}=b/(S+\varepsilon)$.
Its additional numerical perturbation is
\begin{equation}
\xi_{ij}^{\varepsilon}
=|\Lambda_{ij}^{\rm amp}-\lambda_{ij}^{0}|
=\frac{\varepsilon b}{S(S+\varepsilon)}.
\label{eq:app-floor-perturbation}
\end{equation}
Thus
\begin{equation}
|\lambda_{ij}^{\rm act}-\Lambda_{ij}^{\rm amp}|
\leq\delta_j^{\rm rms}+\xi_{ij}^{\varepsilon}.
\label{eq:app-total-proxy-error}
\end{equation}
For fixed $\boldsymbol\rho$ and $\mathbf D$, replacing amplitude shares
by activation-aware shares changes the synthesis target by at most
\begin{equation}
\begin{aligned}
&\|\Delta\mathbf W_{\rm act}^{\star}-\Delta\mathbf W^{\star}\|_F\\
&\quad\leq\|\boldsymbol\rho\odot\mathbf D\odot
(\mathbf1(\boldsymbol\delta^{\rm rms})^\top+
\boldsymbol\Xi^{\varepsilon})\|_F,
\end{aligned}
\label{eq:app-proxy-target-error}
\end{equation}
where $\boldsymbol\delta^{\rm rms}=[\delta_j^{\rm rms}]$,
$\boldsymbol\Xi^{\varepsilon}=[\xi_{ij}^{\varepsilon}]$, and
$\mathbf1\in\mathbb R^m$. With reconstruction residual $e_R$,
\begin{equation}
\begin{aligned}
&\|\widehat{\Delta\mathbf W}-\Delta\mathbf W_{\rm act}^{\star}\|_F\\
&\quad\leq e_R+\|\boldsymbol\rho\odot\mathbf D\odot
(\mathbf1(\boldsymbol\delta^{\rm rms})^\top+
\boldsymbol\Xi^{\varepsilon})\|_F.
\end{aligned}
\label{eq:app-proxy-rank-error}
\end{equation}
The bound connects activation agreement to the synthesis perturbation,
with rank restoration contributing its separate approximation residual.

\subsection{Fixed-Rank Reconstruction}
\label{app:write-proof}

\paragraph{Dense synthesis target}
Hold the synthesis coefficients in Eq. \textcolor{red}{(16)} fixed. For scalar stored/acquired
weights $o,n$, coefficient $\lambda\in[0,1]$, and candidate $w$,
\begin{equation}
\begin{aligned}
&\tfrac12(1-\lambda)(w-o)^2+\tfrac12\lambda(w-n)^2\\
&=\tfrac12\bigl(w-[(1-\lambda)o+\lambda n]\bigr)^2
+\tfrac12\lambda(1-\lambda)(n-o)^2.
\end{aligned}
\label{eq:app-complete-square}
\end{equation}
Summing over entries yields
\begin{equation}
\mathcal J(\mathbf M)
=\tfrac12\|\mathbf M-\Delta\mathbf W^{\star}\|_F^2+C_\Lambda,
\label{eq:app-completed-objective}
\end{equation}
where
\begin{equation}
C_\Lambda=\tfrac12\sum_{a,b}\Lambda_{ab}(1-\Lambda_{ab})D_{ab}^2
\label{eq:app-reconstruction-constant}
\end{equation}
is independent of $\mathbf M$. The Hessian is the identity, so
$\Delta\mathbf W^{\star}$ is the unique dense minimizer.

\paragraph{Rank-constrained optimum}
Let $\mathbf W_\star=\Delta\mathbf W^{\star}$ have singular values
$\sigma_1\geq\cdots\geq\sigma_{d_{\min}}\geq0$. For any
$\operatorname{rank}(\mathbf M)\leq\bar R$, let $\mathbf Q_{\rm col}$
project onto its column space. Orthogonal decomposition gives
\begin{align}
\|\mathbf W_\star-\mathbf M\|_F^2
&\geq\|\mathbf W_\star\|_F^2
-\operatorname{tr}(\mathbf Q_{\rm col}\mathbf W_\star\mathbf W_\star^\top)
\label{eq:app-rank-lower-bound}\\
&\geq\sum_{i>\bar R}\sigma_i^2=e_R^2.
\label{eq:app-svd-tail}
\end{align}
The last inequality follows because the projector's diagonal entries
in the singular-vector basis lie in $[0,1]$ and sum to at most $\bar R$.
Truncated SVD attains equality, and hence minimizes the fixed-coefficient
objective with
\begin{equation}
\mathcal J(\widehat{\Delta\mathbf W})
-\mathcal J(\Delta\mathbf W^{\star})=\tfrac12 e_R^2.
\label{eq:app-objective-gap}
\end{equation}

\paragraph{Factor recovery and displacement}
The recovered factors preserve the original execution scale:
\begin{equation}
\begin{aligned}
\mathbf B^{\star}
&=\varsigma^{-1/2}\mathbf U_{\bar R}\boldsymbol\Sigma_{\bar R}^{1/2},\\
\mathbf A^{\star}
&=\varsigma^{-1/2}\boldsymbol\Sigma_{\bar R}^{1/2}\mathbf V_{\bar R}^{\top},
\end{aligned}
\label{eq:lora-factor-recovery}
\end{equation}
so that
\begin{equation}
\varsigma\mathbf B^{\star}\mathbf A^{\star}
=\mathbf U_{\bar R}\boldsymbol\Sigma_{\bar R}\mathbf V_{\bar R}^{\top}
=\widehat{\Delta\mathbf W}.
\label{eq:app-factor-recovery}
\end{equation}
Unused rows or columns are zero-padded when the prescribed factor
width exceeds the available matrix dimension.

The triangle inequality connects the restored state to its stored state:
\begin{equation}
\|\widehat{\Delta\mathbf W}-\Delta\mathbf W^{\rm old}\|_F
\leq e_R+\|\boldsymbol\Lambda\odot\mathbf D\|_F.
\label{eq:app-restored-displacement}
\end{equation}
Because the stored operator is feasible under the same rank budget,
$e_R\leq\|\boldsymbol\Lambda\odot\mathbf D\|_F$, giving
\begin{equation}
\|\widehat{\Delta\mathbf W}-\Delta\mathbf W^{\rm old}\|_F
\leq2\|\boldsymbol\Lambda\odot\mathbf D\|_F.
\label{eq:app-feasible-old-bound}
\end{equation}
These inequalities separate the retained displacement from the
approximation introduced by rank restoration.

\subsection{Calibration, Retention, and Response Analysis}
\label{app:retention-response}
\label{app:scale-calibration}
\label{app:retention-budget}
\label{app:evidence-proof}
\label{app:stability-proof}
\label{app:historical-risk}

\paragraph{Empirical calibration}

The module ratio $\eta_\ell=m_\ell/\bar z_\ell$ compares the median and
mean squared changes. Rescaling $\mathbf D_\ell$ by $s>0$ gives
\begin{equation}
m'_\ell=s^2m_\ell,\qquad
\bar z'_\ell=s^2\bar z_\ell,\qquad\eta'_\ell=\eta_\ell.
\label{eq:app-scale-invariance}
\end{equation}
Thus each mapping contributes a scale-independent change-profile ratio
to the empirical center, using $\operatorname{Med}_{\rm lo}$ within each
mapping as defined in Eq. \textcolor{red}{(11)}.

For $N=|\mathcal U_t^+|>0$, the equally weighted empirical distribution is
\begin{equation}
\widehat F_t(k)=\frac1N\sum_{\ell\in\mathcal U_t^+}
\mathbb I[\eta_{t,\ell}\leq k].
\label{eq:app-empirical-cdf}
\end{equation}
Order these ratios as $\eta_{(1)}\leq\cdots\leq\eta_{(N)}$.
For $F(k)=\sum_i|\eta_{(i)}-k|$, let $n_<,n_>,n_=$ denote the
counts below, above, and equal to $k$. Then
\begin{equation}
\partial F(k)=[n_<-n_>-n_=,\ n_<-n_>+n_=].
\label{eq:app-median-subgradient}
\end{equation}
The condition $0\in\partial F(k)$ requires at most half the ratios on
either side. Thus the midpoint median $\operatorname{Med}_{\rm mid}$
in Eq. \textcolor{red}{(12)} minimizes $F$; for even $N$, it
selects the midpoint of the minimizing interval.

Substituting $m_\ell=\eta_\ell\bar z_\ell$ gives
\begin{equation}
\tau_\ell^2=\frac{m_\ell}{\kappa_t}
=\frac{\eta_\ell}{\kappa_t}\bar z_\ell.
\label{eq:app-scale-center}
\end{equation}
A mapping at the empirical center uses its mean squared change as
the reference, while other mappings retain their own scale through
$\eta_\ell/\kappa_t$. For all participating mappings, the retention
rule uses the stabilized scale
$\chi_\ell=\max\{\tau_\ell^2,\varepsilon\}+\varepsilon$.

The Gaussian reference in the sensitivity study is a population
quantity. Let $\Phi_{\rm N}$ denote the cumulative distribution
function of the standard normal distribution. For
$d\sim\mathcal N(0,\sigma^2)$ with $\sigma>0$,
$\mathbb E[d^2]=\sigma^2$ and
$\Pr(d^2\leq z)=2\Phi_{\rm N}(\sqrt z/\sigma)-1$ for $z\geq0$.
Setting this probability to one half gives the ratio of the population
median to the mean:
\begin{equation}
\kappa_{\rm G}
=\frac{\operatorname{Med}(d^2)}{\mathbb E[d^2]}
=[\Phi_{\rm N}^{-1}(3/4)]^2\approx0.455.
\label{eq:gaussian-scale-constant}
\end{equation}
ARS instead estimates the center from the participating mappings at
each synthesis step, adapting the reference to the observed change
profiles.

\paragraph{Regularized update retention}
Fix one mapping and omit its index. Let
$\mathbf D^{\rm amp}=\boldsymbol\Lambda^{\rm amp}\odot\mathbf D$,
$\Omega=\{(a,b):Z_{ab}>0\}$, and
$\chi=\max\{\tau^2,\varepsilon\}+\varepsilon$ as in the main text.
For a candidate update $\mathbf Y$, the fidelity and penalty terms in
Eq. \textcolor{red}{(14)} are
\begin{equation}
\begin{aligned}
\mathcal E_{\rm fit}(\mathbf Y)&=\tfrac12\|\mathbf Y-\mathbf D^{\rm amp}\|_F^2,\\
\mathcal E_{\rm pen}(\mathbf Y)&=\tfrac12\sum_{(a,b)\in\Omega}
\frac{\chi}{Z_{ab}}Y_{ab}^2,\qquad
Y_{ab}=0\ \text{outside }\Omega.
\end{aligned}
\label{eq:app-update-fidelity-budget}
\end{equation}
On $\Omega$, the objective has diagonal Hessian entries
$1+\chi/Z_{ab}>0$. Stationarity of the unique minimizer gives
\begin{equation}
\begin{aligned}
\left(1+\frac{\chi}{Z_{ab}}\right)D_{ab}^{\rm ret}&=D_{ab}^{\rm amp},\\
D_{ab}^{\rm ret}&=\frac{Z_{ab}}{Z_{ab}+\chi}D_{ab}^{\rm amp}.
\end{aligned}
\label{eq:app-retention-stationarity}
\end{equation}
Thus $\mathbf D^{\rm ret}=\boldsymbol\rho\odot\mathbf D^{\rm amp}$.
For $z\geq0$ at a fixed $\chi$,
\begin{equation}
\rho(z)=\frac{z}{z+\chi},\ \ 
\rho'(z)=\frac{\chi}{(z+\chi)^2}>0,\ \ 
0\leq\rho(z)<1.
\label{eq:app-confidence-properties}
\end{equation}
Stronger relative changes retain a larger fraction of the proposal,
while weaker ones receive stronger regularization. Entry-wise bounds
also give
\begin{equation}
\begin{aligned}
\|\boldsymbol\rho\odot\boldsymbol\Lambda^{\rm amp}\odot\mathbf D\|_F^2
&\leq\|\boldsymbol\Lambda^{\rm amp}\odot\mathbf D\|_F^2\\
&\leq\|\mathbf D\|_F^2.
\end{aligned}
\label{eq:app-write-contraction}
\end{equation}
Combining Eq.~(\ref{eq:app-write-contraction}) with
Eq.~(\ref{eq:app-restored-displacement}) proves
Eq. \textcolor{red}{(19)} and connects change-dependent retention
to the final fixed-rank displacement.

\paragraph{Response-loss variation}
For a fixed module input $\mathbf h$, let $\delta\mathbf W$ denote
the perturbation of the module's effective LoRA weight. Then
\begin{equation}
\|\delta\mathbf W\mathbf h\|_2
\leq\|\delta\mathbf W\|_F\|\mathbf h\|_2.
\label{eq:app-module-output}
\end{equation}
Writing $b_\ell=\|\boldsymbol\Lambda_\ell\odot\mathbf D_\ell\|_F+e_{R,\ell}$,
the restored mapping therefore satisfies
\begin{equation}
\|(\widehat{\Delta\mathbf W}_\ell-
\Delta\mathbf W_\ell^{\rm old})\mathbf h\|_2
\leq b_\ell\|\mathbf h\|_2.
\label{eq:local-output-bound}
\end{equation}
To analyze the complete response, let $\boldsymbol\vartheta$ collect
the reused root's vectorized effective LoRA matrices and projector
parameters. Its dimension is $D_{\rm eff}+D_{\rm proj}$, where
$D_{\rm eff}=\sum_\ell D_{{\rm out},\ell}D_{{\rm in},\ell}$.

\begin{assumption}[Local response-loss regularity]
\label{ass:local-response}
With inference probabilities fixed, the response loss is differentiable
along $\boldsymbol\vartheta(s)=\boldsymbol\vartheta_o+
s(\boldsymbol\vartheta_f-\boldsymbol\vartheta_o)$, $s\in[0,1]$,
and its block gradients satisfy
\begin{equation}
\begin{aligned}
\|\nabla_{\Delta\mathbf W_\ell}\ell_{\rm resp}(\boldsymbol\vartheta(s);\xi)\|_F
&\leq \Gamma_\ell(\xi),\\
\|\nabla_{\boldsymbol\Phi}\ell_{\rm resp}(\boldsymbol\vartheta(s);\xi)\|_2
&\leq \Gamma_\Phi(\xi).
\end{aligned}
\label{eq:app-local-smoothness}
\end{equation}
\end{assumption}

Integrating the directional derivative and applying Cauchy--Schwarz
to each block gives
\begin{equation}
\begin{aligned}
|\ell_{\rm resp}(\boldsymbol\vartheta_f;\xi)
-\ell_{\rm resp}(\boldsymbol\vartheta_o;\xi)|
\leq\sum_\ell \Gamma_\ell(\xi)b_\ell+\Gamma_\Phi(\xi)d_\Phi,
\end{aligned}
\label{eq:app-loss-stability}
\end{equation}
where $d_\Phi=\|\boldsymbol\Phi^{\rm new}-\boldsymbol\Phi^{\rm old}\|_2$.
Thus, response variation depends jointly on the retained LoRA update,
its rank-restoration residual, and projector adaptation.

Let $\mathbb E_{\rm old}$ denote expectation under a fixed mixture
of earlier-task distributions. The historical risk is
\begin{equation}
\mathcal R_{\rm old}(\boldsymbol\vartheta)
=\mathbb E_{\rm old}\!\left[\ell_{\rm resp}(\boldsymbol\vartheta;\xi)\right].
\label{eq:app-historical-risk-definition}
\end{equation}
For the $\beta$-Lipschitz gradient in
Proposition \textcolor{red}{1}, write
$\boldsymbol\delta=\boldsymbol\vartheta_f-\boldsymbol\vartheta_o$ and
$\mathbf g_o=\nabla\mathcal R_{\rm old}(\boldsymbol\vartheta_o)$.
Integration yields
\begin{equation}
\begin{aligned}
\Delta\mathcal R_{\rm old}
&=\int_0^1\langle\nabla\mathcal R_{\rm old}
(\boldsymbol\vartheta_o+s\boldsymbol\delta),\boldsymbol\delta\rangle\,ds\\
&\leq\langle\mathbf g_o,\boldsymbol\delta\rangle
+\frac\beta2\|\boldsymbol\delta\|_2^2.
\end{aligned}
\label{eq:app-smooth-historical-risk}
\end{equation}
Applying the block norm bounds proves
Eq. \textcolor{red}{(20)}; at a stationary stored state,
the first-order terms vanish. 
We compare ARS and a reference update $c$ under the same historical
risk $\mathcal R_{\rm old}$, with the other experts and root routing
weights held fixed across both evaluations. Both updates start from
the same stored state and use the same final projector. Assuming a
common gradient Lipschitz constant $\beta$ along the two update
segments, the upper bound on the increase in historical risk is no
larger for ARS whenever
\begin{equation}
\|\mathbf D_\ell^{\rm ret}\|_F+e_{R,\ell}
\leq\|\mathbf D_\ell^{\rm cmp}\|_F+e_{R,\ell}^{\rm cmp}
\quad\text{for all }\ell,
\label{eq:app-risk-bound-comparison}
\end{equation}
where $\mathbf D_\ell^{\rm cmp}$ and $e_{R,\ell}^{\rm cmp}$ are the
comparator's retained update and rank-restoration residual. This follows
from the monotonicity of $G_\ell b+\beta b^2/2$ over $b\geq0$.

When a new identity reuses root $r_k$, the existing root set is unchanged.
For a fixed instruction, let $\pi_{\rm new}^{\rm task}$ be the new
identity's routing weight after insertion and write
$\boldsymbol\pi_o^{\rm root}$ and $\boldsymbol\pi_f^{\rm root}$
for the root-weight vectors before and after insertion. Since the
historical scores remain fixed, each old identity weight is multiplied
by $1-\pi_{\rm new}^{\rm task}$. Consequently,
\begin{equation}
\begin{aligned}
\boldsymbol\pi_f^{\rm root}
&=(1-\pi_{\rm new}^{\rm task})\boldsymbol\pi_o^{\rm root}
+\pi_{\rm new}^{\rm task}\mathbf e_k,\\
\|\boldsymbol\pi_f^{\rm root}-\boldsymbol\pi_o^{\rm root}\|_1
&=2\pi_{\rm new}^{\rm task}(1-\pi_{o,k}^{\rm root}).
\end{aligned}
\label{eq:app-routing-transition}
\end{equation}
Here $\mathbf e_k$ is the unit vector for the reused root. The change
in expert weights is small when the reused root already carries most
of the mass.

The complete response transition can be separated as
\begin{equation}
\begin{aligned}
\Delta\ell_{\rm resp}={}&
\ell_{\rm resp}(\boldsymbol\vartheta_f,\boldsymbol\pi_o^{\rm root};\xi)
-\ell_{\rm resp}(\boldsymbol\vartheta_o,\boldsymbol\pi_o^{\rm root};\xi)\\
&+\ell_{\rm resp}(\boldsymbol\vartheta_f,\boldsymbol\pi_f^{\rm root};\xi)
-\ell_{\rm resp}(\boldsymbol\vartheta_f,\boldsymbol\pi_o^{\rm root};\xi).
\end{aligned}
\label{eq:app-full-transition}
\end{equation}
If $\|\nabla_{\boldsymbol\pi^{\rm root}}\ell_{\rm resp}\|_\infty\leq \Gamma_\pi(\xi)$
along the probability segment at the final parameter state, integrating
this second segment and using Eq.~(\ref{eq:app-loss-stability}) gives
\begin{equation}
\begin{aligned}
|\Delta\ell_{\rm resp}|&\leq\sum_\ell \Gamma_\ell(\xi)b_\ell+\Gamma_\Phi(\xi)d_\Phi\\
&\quad+\Gamma_\pi(\xi)\|\boldsymbol\pi_f^{\rm root}-\boldsymbol\pi_o^{\rm root}\|_1.
\end{aligned}
\label{eq:app-full-response-bound}
\end{equation}
The two adaptation paths and their common routing distribution thus
contribute explicitly to response variation. The following analysis
examines how these components affect acquisition and retention during
expert reuse.

\subsection{Analysis of Shared-Expert Adaptation}
\label{app:shared-adaptation}
\label{app:projector-proof}

\paragraph{LoRA synthesis}
We examine the reuse of the VizWiz expert when Flickr30k arrives in
UCIT. The stored LoRA preserves its pre-acquisition state, direct
adoption uses the newly trained state, and amplitude-only synthesis
weights their difference by relative amplitude. ARS additionally
regulates this proposed update with change confidence. Each condition
uses the same acquired projector and pre-acquisition routing weights.
All synthesized targets are restored under the same LoRA rank budget.
For evaluation example $n$ of task $a$, let
$(\mathbf X_{a,n}^{\rm eval},\mathbf u_{a,n}^{\rm eval},
\mathbf y_{a,n}^{\rm eval})$ denote the image, instruction, and target
response, and let $N_{a,n}^{\rm ans,eval}$ be the target-token count.
The token distribution $p_{\mathcal M}$ is now evaluated under the
projector, LoRA, and routing configuration being compared. The
task-level response NLL is
\begin{equation}
\begin{aligned}
\mathcal L_a^{\rm eval}=-&
\frac{1}{\sum_{n=1}^{N_a^{\rm eval}} N_{a,n}^{\rm ans,eval}}
\sum_{n=1}^{N_a^{\rm eval}}\sum_{q=1}^{N_{a,n}^{\rm ans,eval}}\\
&\log p_{\mathcal M}\!\left(y_{a,n,q}^{\rm eval}
\mid \mathbf X_{a,n}^{\rm eval},\mathbf u_{a,n}^{\rm eval},
\mathbf y_{a,n,<q}^{\rm eval}\right).
\end{aligned}
\label{eq:app-evaluation-nll}
\end{equation}
The earlier-task average weights the five preceding tasks equally.
\begin{table}[!t]
\centering
\caption{Response NLL of alternative LoRA synthesis rules at UCIT expert
reuse. All conditions use the same acquired projector and pre-acquisition
routing. Earlier Avg. equally weights the five preceding tasks.}
\label{tab:retention-nll}
\begingroup
\small
\setlength{\tabcolsep}{4pt}
\renewcommand{\arraystretch}{1.08}
\begin{tabular}{@{}lccc@{}}
\toprule
LoRA synthesis & Earlier Avg.$\downarrow$
& VizWiz$\downarrow$ & Flickr30k$\downarrow$ \\
\midrule
Stored LoRA & 0.523975 & 1.814302 & 1.811942 \\
Direct adoption & 0.561092 & 1.999998 & 1.757610 \\
Amplitude only & 0.542216 & 1.905882 & 1.756630 \\
\rowcolor{gray!10}
ARS & 0.534833 & 1.868450 & 1.762335 \\
\bottomrule
\end{tabular}
\endgroup
\end{table}

Table~\ref{tab:retention-nll} shows that direct adoption improves
the arriving task at the expense of earlier responses. Amplitude-only
synthesis alleviates this interference, and ARS further reduces the
VizWiz loss increase from $0.091580$ to $0.054148$, a $40.87\%$
reduction, while retaining $89.68\%$ of the Flickr30k NLL improvement.
The mean loss increase over earlier tasks also decreases. These results
support combining amplitude weighting with change-dependent retention
to preserve most current-task acquisition while moderating interference
with the shared expert's earlier capabilities.

\paragraph{Continued projector adaptation}
For frozen visual tokens, the two-layer projector is
\begin{equation}
\begin{aligned}
\mathcal P_g(\mathbf X^{\rm vis};\boldsymbol\Phi_g)
={}&\operatorname{GELU}(\mathbf X^{\rm vis}\mathbf W_{1,g}
+\mathbf1\mathbf b_{1,g}^{\top})\mathbf W_{2,g}
+\mathbf1\mathbf b_{2,g}^{\top},
\end{aligned}
\label{eq:projector}
\end{equation}
where $\mathbf W_{1,g}\in\mathbb R^{D_v\times D_m}$,
$\mathbf W_{2,g}\in\mathbb R^{D_m\times D_m}$, and both biases
have dimension $D_m$. Parameter interpolation generally differs from
output interpolation. For a scalar restriction with zero biases and
second-layer weight one, interpolating first-layer preactivations zero
and two at coefficient $1/2$ yields $\operatorname{GELU}(1)$;
interpolating their outputs yields $\tfrac12\operatorname{GELU}(2)$.
For exact GELU these equal $\Phi_{\rm N}(1)$ and $\Phi_{\rm N}(2)$,
respectively. \method{} therefore treats projector adaptation separately
from effective LoRA synthesis and combines projector outputs at inference.

From the same five-task UCIT state, we adapt Flickr30k with the selected
projector either trainable or frozen during reuse. Each branch continues
LoRA training and applies ARS, keeping the training seed, data order,
initial state, and evaluation examples fixed. Within each branch, we
evaluate the $3\times2\times3$ combinations of old/direct/ARS LoRA,
old/new projector, and old/new/oracle root probabilities. The final ARS
models use their acquired projectors and updated root probabilities;
we also compare stored/acquired projector outputs on the same images.
Captioning scores average BLEU-1--4, METEOR, ROUGE-L, and CIDEr.
Sample-level uncertainty is estimated with 2,000 paired bootstrap
replicates.
\begin{figure*}[!t]
\centering
\subfloat[Trainable projector during reuse]{
\includegraphics[width=.96\textwidth]{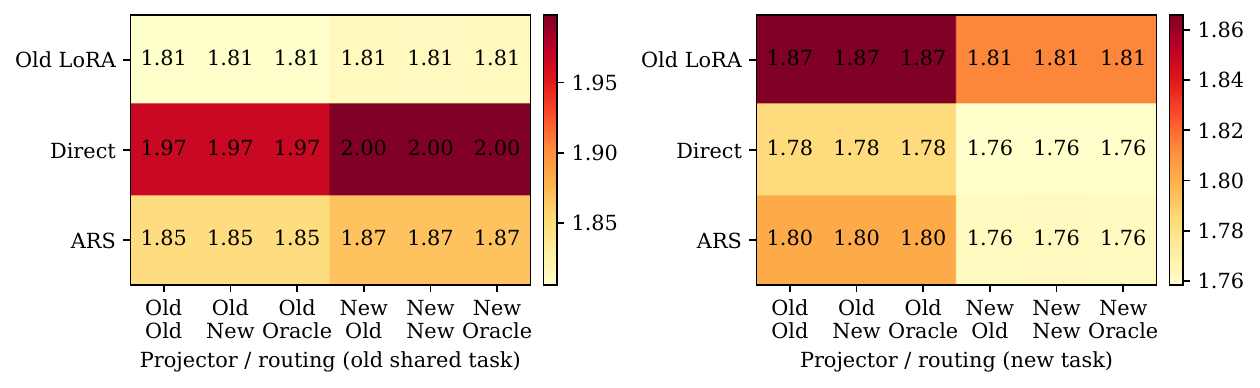}}\\[-1pt]
\subfloat[Frozen projector during reuse]{
\includegraphics[width=.96\textwidth]{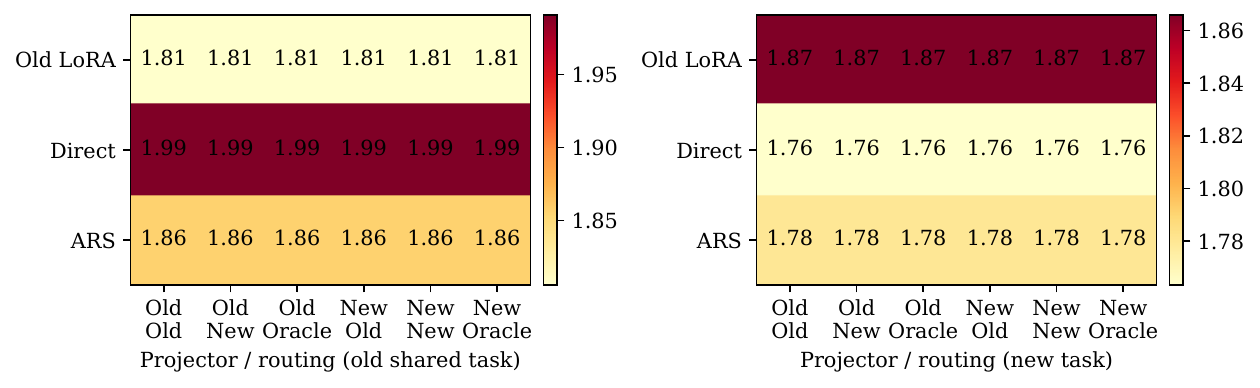}}
\caption{Response NLL under different LoRA, projector, and root-probability
states. Rows specify old, directly acquired, or ARS LoRA weights.
Columns specify the projector state (upper label) and root probabilities
(lower label). The left and right panels evaluate VizWiz and Flickr30k,
respectively. In (b), the projector is fixed throughout reuse.
Oracle probabilities select the task's assigned root.}
\label{fig:q2-projector-factorial}
\end{figure*}
\begin{table}[!t]
\centering
\caption{Continued projector adaptation on the shared UCIT expert.
Response NLL compares frozen and trainable projector branches under ARS.
Output cosine compares the stored and acquired projector on the same
images in the trainable branch.}
\label{tab:projector-nll-cosine}
\begingroup
\small
\setlength{\tabcolsep}{4pt}
\renewcommand{\arraystretch}{1.08}
\begin{tabular}{@{}lccc@{}}
\toprule
& \multicolumn{2}{c}{Response NLL$\downarrow$}
& Output \\
\cmidrule(lr){2-3}
Task & Frozen & Trainable & cosine \\
\midrule
VizWiz & 1.855822 & 1.866667 & 0.967923 \\
Flickr30k & 1.779504 & 1.762124 & 0.964773 \\
\bottomrule
\end{tabular}
\endgroup
\end{table}

Table~\ref{tab:projector-nll-cosine} combines response NLL with the
similarity of the old and adapted projector outputs on the same images.
Training the projector decreases Flickr30k NLL by $0.017380$
(paired 95\% interval $[-0.027809,-0.007489]$) and increases
VizWiz NLL by $0.010845$ (interval $[0.001183,0.020278]$).
The corresponding caption scores change from $58.878$ to $61.183$
on Flickr30k and from $56.320$ to $55.245$ on VizWiz, increasing their
mean from $57.599$ to $58.214$. Output cosines of $0.967923$ and
$0.964773$ show that the adapted mapping remains directionally aligned
with the old mapping. Continued projector adaptation therefore improves
current-task alignment while maintaining a favorable task-pair average,
complementing the LoRA retention examined above.

Fig.~\ref{fig:q2-projector-factorial} distinguishes the roles of projector
adaptation and LoRA synthesis. In the trainable branch, ARS reduces the
VizWiz loss increase from $0.192378$ under direct adoption to $0.060967$,
while retaining $96.47\%$ of Flickr30k's NLL improvement relative to
the complete pre-acquisition state. Comparing the old and acquired
projector columns shows that visual alignment supports new-task
acquisition, while ARS moderates changes in the shared LoRA state.

\paragraph{Root-level inference}

Within each projector condition in
Fig.~\ref{fig:q2-projector-factorial}, the old, updated, and oracle root
probabilities yield closely matched NLLs on the shared tasks. With ARS
and the acquired projector fixed, replacing old by updated root
probabilities changes NLL by $0.000053$ on VizWiz and $0.000055$ on
Flickr30k, with paired intervals containing zero. Equation~(\ref{eq:app-routing-transition})
explains this stability: both task identities contribute to the same
expert's aggregated probability. Root-level execution thus preserves
stable expert weighting while the shared adaptation state incorporates
the arriving task.

Together, these analyses support retaining the supervised projector for
current-task alignment and using ARS to control shared LoRA updates.
The projector term and routing term in
Eq.~(\ref{eq:app-full-response-bound}) correspond to distinct components
of this acquisition--retention balance, complementing the retained-update
and rank-restoration terms.

\vfill

\end{document}